\pdfoutput=1

\documentclass[11pt]{article}

\usepackage[final]{acl}

\usepackage{times}
\usepackage{latexsym}

\usepackage[T1]{fontenc}

\usepackage[utf8]{inputenc}

\usepackage{microtype}

\usepackage{inconsolata}

\usepackage{graphicx}
\usepackage{enumitem}
\usepackage{listings}
\usepackage{booktabs}
\usepackage{multicol}
\usepackage{multirow}
\usepackage{graphicx}
\usepackage{tabularx}
\usepackage{arydshln}
\usepackage{float}
\usepackage{placeins}
\usepackage{xcolor}
\usepackage[hang,flushmargin]{footmisc}

\title{TopXGen: Topic-Diverse Parallel Data Generation \\for Low-Resource Machine Translation}

\author{Armel Zebaze 
\quad Benoît Sagot\quad Rachel Bawden\\
Inria, Paris, France\\
\texttt{firstname.lastname@inria.fr}\\}

\begin{document}
\maketitle
\begin{abstract}

LLMs have been shown to perform well in machine translation (MT) with the use of in-context learning (ICL), rivaling supervised models when translating into high-resource languages (HRLs). However, they lag behind when translating into low-resource language (LRLs).
Example selection via similarity search and supervised fine-tuning help. However the improvements they give are limited by the size, quality and diversity of existing parallel datasets. 
A common technique in low-resource MT is synthetic parallel data creation, the most frequent of which is backtranslation, whereby existing target-side texts are automatically translated into the source language. However, this assumes the existence of good quality and relevant target-side texts, which are not readily available for many LRLs.
In this paper, we present \textsc{TopXGen}, an LLM-based approach for the generation of high quality and topic-diverse data in multiple LRLs, which can then be backtranslated to produce useful and diverse parallel texts for ICL and fine-tuning.
Our intuition is that while LLMs struggle to translate into LRLs, their ability to translate well into HRLs and their multilinguality enable them to generate good quality, natural-sounding target-side texts, which can be translated well into a high-resource source language.
We show that \textsc{TopXGen} boosts LLM translation performance during fine-tuning and in-context learning.
%
Code and outputs are available at \url{https://github.com/ArmelRandy/topxgen}.
\end{abstract}

\begin{figure}[ht]
\begin{center}
    \includegraphics[width=0.92\linewidth]{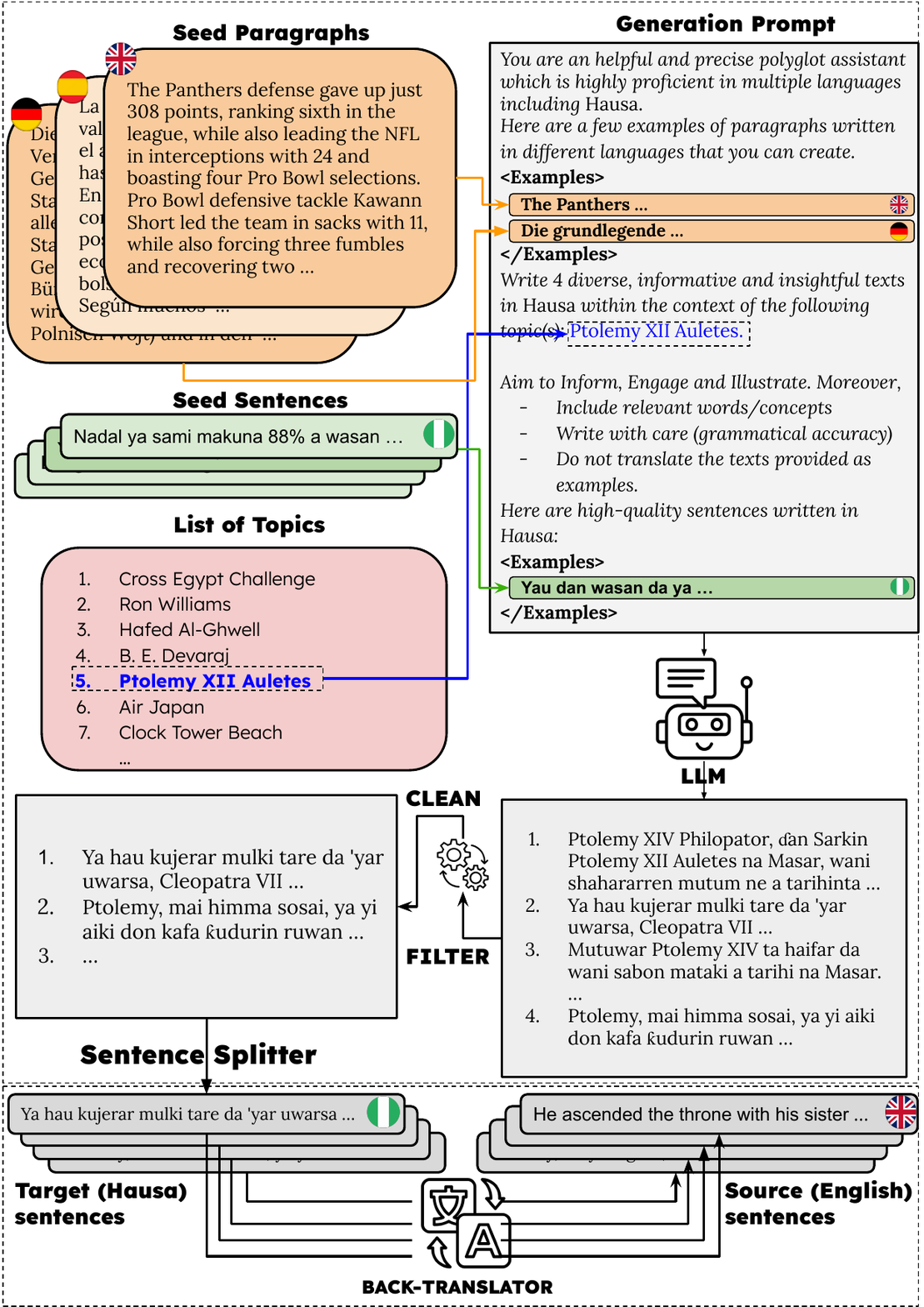}
    \caption{Overview of \textsc{TopXGen}. An LLM is used to write a diverse set of paragraphs in an LRL guided by topics, example sentences in the LRL and example paragraphs in HRLs. The generated paragraphs are later cleaned and divided into sentences that are back-translated into the source language to create a sentence-level parallel dataset.}
    \label{fig:fig1}
\end{center}
\end{figure}

\section{Introduction}

The performance of Machine Translation (MT) models has considerably evolved through the years, with new models increasingly supporting more languages~\citep{bapna2022buildingmachinetranslationsystems, nllb2022, yang2023bigtranslateaugmentinglargelanguage}. However, performance remains unequal across languages, which themselves vary greatly in terms of available resources and representation in NLP research \citep{joshi-etal-2020-state}. MT Models perform much better in high-resource languages (HRLs), such as English, French and German, compared to low-resource languages (LRLs) such as Hausa, which lack large quantities of high-quality parallel data. 
Decoder-based LLMs can perform MT without relying on parallel data (in a zero-shot fashion). However they lag behind supervised MT baselines when translating into LRLs \citep{hendy2023goodgptmodelsmachine}. In-Context Learning (ICL; \citealp{NEURIPS2020_1457c0d6}), which involves using a few high-quality demonstrations has been shown to improve performance, especially when they are similar to the sentence to be translated~\citep{moslem-etal-2023-adaptive, zebaze-etal-2025-context}, highlighting the importance of parallel datasets even in this setting.
A common approach is to synthesize parallel data using forward translation or back-translation \citep{schwenk-2008-investigations, bojar-tamchyna-2011-improving, sennrich-etal-2016-improving}.
Forward translation comes with issues such as low translation fidelity and the neglect of cultural nuances in the target language. 
Back-Translation (BT; \citealp{sennrich-etal-2016-improving}) typically relies on a good quality monolingual corpus in the LRL, which can be difficult to obtain. 
In this work, we explore the construction of synthetic datasets for MT into LRLs by automatically generating text in the LRL target language and then backtranslating into HRL source language. 
\citet{marie2021synthesizingmonolingualdataneural} first proposed an approach with a similar aim, using an LLM to generate in-domain monolingual data, which they then used to perform BT alongside parallel data. They generate their monolingual data using an LLM fine-tuned on each of their domain-specific datasets. 
However, their approach focuses on English and requires domain-specific fine-tuning (and datasets) in addition to relying on parallel data to perform BT, making it impractical for translation into LRLs. 
To address this, we introduce \textsc{TopXGen} (Figure~\ref{fig:fig1}), a pipeline that exploits LLMs' multilinguality and instruction-following capabilities. Unlike prior work, we generate monolingual data (sentences) beyond English to cover numerous LRLs. Instead of domain-specific fine-tuning, we directly prompt an LLM and guide its generations towards a predefined list of topics in order to encourage diversity in the outputs. 
The quality of sentences stems from the ability of state-of-the-art multilingual LLMs to produce coherent text in LRLs \citep{enis2024llmnmtadvancinglowresource}---even if they struggle to translate accurately into them. We then backtranslate the generated sentence into a HRL (English in this work) using a reverse translation model (a supervised MT system or the same LLM). 
This target-aware generation helps mitigate the cultural loss often observed in LRLs with standard forward translation techniques. Moreover, translating into a HRL offers a practical advantage, as HRLs are generally easier to translate into with high fidelity. 

To evaluate \textsc{TopXGen}, we generate synthetic parallel data between English and ten low-resource languages—Basque, Hausa, Igbo, Kinyarwanda, Nepali, Somali, Sundanese, Swahili, Urdu, and Xhosa—using \texttt{Gemma-3-27B-It} as the generator and \texttt{NLLB-200-3.3B} as the back-translator. We then assess both in-context learning and fine-tuning setups across small translation models. Our experiments show that \textsc{TopXGen} consistently outperforms other data generation methods and achieves performance comparable to human-translated datasets. 

\section{Related Work}

\paragraph{Low-resource Machine Translation with LLMs.}
LLMs encounter many languages during their training but in various proportions~\citep{abadji-etal-2022-towards, penedo2024fineweb-2}. Through ICL \citep{NEURIPS2020_1457c0d6}, they can perform a wide variety of tasks including MT. Decoder-based LMs are on par with supervised MT models ---such as M2M100 \citep{fan2021beyond} and NLLB \citep{goyal-etal-2022-flores}---when translating between high-resource languages but still lag behind when translating into low-resource languages \citep{hendy2023goodgptmodelsmachine, zhu-etal-2024-multilingual}. Many works have emerged to bridge the gap at inference time by either using similarity-based in-context example selection \citep{moslem-etal-2023-adaptive, tanzer2024a, zebaze-etal-2025-context} or more advanced prompting strategies~\citep{he2024exploring, briakou-EtAl:2024:WMT, zebaze2025compositionaltranslationnovelllmbased}. Another line of work involves fine-tuning LLMs. While most works focus on mid- to high-resource languages~\citep{xu2024a, xu2024contrastive, xu2025xalma}, a few of them explore fine-tuning on LRLs. They generally use bitexts mined from the internet \citep{schwenk-etal-2021-wikimatrix, el-kishky-etal-2020-ccaligned, schwenk-etal-2021-ccmatrix}, out-of-domain benchmarks written by natives in low-resource languages~\citep{muennighoff-etal-2023-crosslingual, ustun-etal-2024-aya, uemura-etal-2024-afriinstruct} or continual pretraining on monolingual data covering multiple LRLs to improve few-shot MT performance~\citep{buzaaba2025lughallamaadaptinglargelanguage}. The scarcity of high-quality parallel data is the major bottleneck for low-resource MT but some studies have explored methods to generate such data using LLMs.

\paragraph{Parallel Data Generation.}
Generating data using LLMs has emerged as a popular alternative to costly human annotation, primarily for instruction datasets. One of the first approaches to demonstrate the effectiveness of this paradigm is \textsc{Self-Instruct} \citep{wang-etal-2023-self-instruct, alpaca}. It consists in bootstrapping a small set of seed instructions into a bigger collection with the help of ICL. Building on this, \citet{kou-etal-2024-knn} proposed \textsc{KNN-Instruct} which replaces the random selection of ICL demonstrations with nearest neighbor retrieval in an embedding space. Subsequent advances introduced multilingual instruction generation \citep{cui2024efficienteffectivetextencoding, wei2023polylmopensourcepolyglot}, generating increasingly complex and diverse instructions across domains  \citep{xu2024wizardlm, zeng-etal-2024-automatic, codealpaca, luo2025wizardmath, luo2024wizardcoder} and step-by-step explanations within responses \citep{mukherjee2023orcaprogressivelearningcomplex, gunasekar2023textbooksneed, li2023textbooksneediiphi15}. Other approaches leverage unlabeled human-written corpora as sources for generating instruction responses \citep{wei2024magicoder, wei2024selfcodealign, li2024selfalignment, benallal2024cosmopedia}.
For MT and LRLs, many works generate datasets by simply machine translating existing ones in the languages of interest (The Aya Collection; \citealp{singh-etal-2024-aya}, the XLLMs-100 collection \citealp{lai-etal-2024-llms} and Bactrian-X; \citealp{li2023bactrianxmultilingualreplicableinstructionfollowing}).
One of the most common and early strategies was to backtranslate monolingual target side data into the source language to create synthetic parallel data \citep{schwenk-2008-investigations, bertoldi-federico-2009-domain, bojar-tamchyna-2011-improving,sennrich-etal-2016-improving,caswell-etal-2019-tagged,burlot-yvon-2018-using, bogoychev2020domaintranslationesenoisesynthetic, marie-etal-2020-tagged}. 
However, BT's reliance on high-quality monolingual data on the target side presents a challenge when such data is hard to obtain. To address this, \citet{marie2021synthesizingmonolingualdataneural} proposed using LLMs to synthesize monolingual data by fine-tuning GPT-2 \citep{radford2019language} on multiple domains and generating multi-domain English data. Combined with substantial amounts of parallel data, they apply BT to train a model to translate from a target language into English. We propose to also generate synthetic parallel datasets, but for low-resource MT instead of HRL translation for multi-domain adaptation, the challenge being to be able to generate high quality data in a LRL. Instead of fine-tuning an LLM, our approach uses more recent strategies involving LLM prompting to generate high quality and diverse data, which we then use to perform ICL and fine-tuning for translation into LRLs.

\section{Methodology}

\begin{table*}[ht]
\captionsetup{skip=0pt}
\small
\begin{center}
\resizebox{\linewidth}{!}{
\begin{tabular}{lrrrrrrrrrr}
\toprule
{} & Basque & Hausa & Igbo & Kinyarwanda & Nepali & Somali & Sundanese & Swahili & Urdu & Xhosa \\
\midrule
Paragraphs                       & 16,829 & 14,981 & 18,518 & 8,900 & 14,490 & 14,623 & 10,483 & 11,489 & 13,923 & 15,781 \\
Sentences                        & 120,031 & 101,488 & 133,071 & 57,884 & 143,014 & 96,315 & 78,264 & 86,981 & 131,133 & 104,992 \\
Sentences (\small{After decon.}) & 120,031 & 101,466 & 133,063 & 57,884 & 142,681 & 96,315 & 78,257 & 86,981 & 131,118 & 104,979 \\
\bottomrule
\end{tabular}
}
\end{center}
\caption{Statistics of the \textsc{TopXGen} dataset in terms of paragraphs and sentences.}
\label{tab:statistics}
\end{table*}

We propose \textsc{TopXGen} (Figure~\ref{fig:fig1}), a topic-guided and target-language centric method for automatically generating parallel sentence-level datasets between English and LRLs, with the end goal of improving MT into LRLs via ICL or supervised fine-tuning using the synthetic examples. It consists of two steps: data generation and back-translation.

\paragraph{Data generation.}
We generate data in the LRL of interest by prompting a multilingual LLM. We aim to produce data that is structurally and lexically diverse by generating it at the paragraph level data (which we then split into sentences before backtranslation) and by guiding the generation with predefined topics. To generate a new paragraph in a given target LRL, we prompt the LLM with:
\begin{itemize}
    \item \textbf{Topics:} To foster diversity in the output texts. We rely on the generator's instruction-following abilities and prompt it to generate content on a randomly selected topic drawn from a predefined list of Wikipedia topics \cite{mwcat} following \citep{li2023textbooksneediiphi15}.\footnote{In contrast to \citet{marie2021synthesizingmonolingualdataneural}, who focus on broad domains like IT or Health, we use fine-grained topics (e.g.,~specific personalities or events).}
    \item \textbf{Seed paragraphs:} To have the generator understand what we expect in terms of length and format. We use the 240 paragraphs from XQuAD~\citep{artetxe-etal-2020-cross}, which are written in 11 HRLs and perform cross-lingual ICL (X-ICL; \citealp{cahyawijaya-etal-2024-llms}).
    \item \textbf{Seed sentences.} These serve to illustrate how a sentence is structured in the LRL and to help ensure the outputs are generated in the correct script. We use the FLORES-200 dev set \citep{goyal-etal-2022-flores, nllb2022}.
\end{itemize}
Note that during generation, we automatically remove generations that overlap too much with previous ones with respect to the ROUGE\footnote{Version: 0.1.2$|$Pure python implementation of ROUGE-1.5.5} score~\citep{lin-2004-rouge} following \citep{wang-etal-2023-self-instruct}.
Given the collection of paragraphs produced by the generator, we build a collection of sentences by applying a sentence-splitter.\footnote{\url{https://github.com/mediacloud/sentence-splitter}} We then perform language identification with fastText~\citep{bojanowski-etal-2017-enriching, nllb2022} on each sentence and remove incorrectly labeled ones. 
\paragraph{Back-Translation.}
We then use a multilingual back-translator (e.g., \texttt{NLLB-200-3.3B}; \citealp{nllb2022}) to translate the generated sentences into the HRL we want to learn to translate from. Translating into the HRL is likely to be of good quality given that MT models perform better in this direction than the reverse one. 

\subsection{The \textsc{TopXGen} dataset}

We generate multiple paragraphs in 10 languages: 
Basque, Hausa, Igbo, Kinyarwanda, Nepali, Somali, Sundanese, Swahili, Urdu and Xhosa. 
Following \citet{gunasekar2023textbooksneed} and \citet{benallal2024cosmopedia}, to avoid data contamination, we filtered the generated paragraphs to remove those containing a significant overlap (at least one 10-gram overlap) with FLORES (also NTREX~128 and TICO-19, whose results are given in Appendix~\ref{appendix:ntrex_and_tico}). We apply the same strategy to the English translations of the sentences to make sure that they do not resemble the seed paragraphs that we used (i.e.~XQuAD; \citealp{artetxe-etal-2020-cross}).
In Table~\ref{tab:statistics}, we report the number of paragraphs generated per language and the number of sentences before and after decontamination. Each language has between 50k and 150k sentences for a total of 1.05M sentences. We conduct further analysis in 
Appendices~\ref{appendix:dataset}, ~\ref{appendix:topic_modeling} and~\ref{appendix:qualitative_analysis}.

\section{Experiments}

\begin{table*}[ht]
\captionsetup{skip=0pt}
\vskip 0.15in
\small
\begin{center}
\resizebox{\textwidth}{!}{
\begin{tabular}{lrrrrrrrrrrrrrr}
\toprule
\multirow{2}{*}{Models}  & \multicolumn{2}{c}{Basque} & & \multicolumn{2}{c}{Hausa} & & \multicolumn{2}{c}{Igbo} & & \multicolumn{2}{c}{Kinyarwanda} & & \multicolumn{2}{c}{Nepali}\\
\cmidrule{2-3} \cmidrule{5-6} \cmidrule{8-9} \cmidrule{11-12} \cmidrule{14-15}
{} & {BLEU} & {MetricX} &  & {BLEU} & {MetricX} & & {BLEU} & {MetricX} & & {BLEU} & {MetricX} & & {BLEU} & {MetricX}\\
\midrule
{} & \multicolumn{14}{c}{\textit{Toplines}} \\
\midrule
\texttt{NLLB-200-3.3B}       & 24.96 & 6.11 &  & 28.82 & 2.46 & & 19.92 & 4.89 & & 23.35 & 4.58 & & 26.64 & 6.60 \\
\texttt{Gemma-3-27B-It}      & 27.17 & 4.81 &  & 19.03 & 4.84 & & 15.37 & 7.56 & & 12.61 & 
7.92 & & 25.74 & 3.47 \\
\midrule
{} & \multicolumn{14}{c}{\textit{Baselines}} \\
\midrule
\texttt{Gemma-2-27B-It}      & 23.33 & 6.21 &  & 17.54 & 5.54 & & 11.71 & 10.89 & & 7.00 & 14.83 & & 22.63 & 4.00 \\
\texttt{LLaMA-3.1-70B It}    & 26.06 & 5.15 &  & 19.01 & 5.90 & & 15.59 & 8.37 & & 8.25 & 13.45 & & 25.33 & 4.31 \\
\texttt{Gemma-2-9B-It}       & 16.69 & 9.33 &  & 13.76 & 7.13 & & 9.06 & 14.65 & & 4.33 & 19.99 & & 18.91 & 4.80 \\
\texttt{Command-R7B}         & 3.16 & 13.34 &  & 1.88 & 20.37 & & 2.12 & 21.46 & & 2.19 & 22.60 & & 5.37 & 9.12 \\
\texttt{Aya-expanse-32B}     & 8.35 & 17.06 &  & 4.73 & 17.41 & & 4.51 & 21.35 & & 3.51 & 21.18 & & 9.96 & 7.91 \\
\texttt{Qwen-2.5-32B-It}     & 7.27 & 18.99 &  & 4.26 & 16.68 & & 5.76 & 20.03 & & 2.99 & 22.84 & & 9.86 & 9.26 \\
\midrule
\texttt{LLaMAX3-8B Alpaca}      & 12.14 & 10.57 &  & 17.50 & 6.33 & & 13.57 & 9.34 & & 4.06 & 19.29 & & 21.20 & 5.31 \\
\midrule
\texttt{LLaMA-2-7B}  \small{\textsc{5-shot Bm25}}        & 3.42 & 23.02 &  & 1.71 & 22.16 & & 2.03 & 23.08 & & 2.58 & 13.92 & & 3.22 & 15.40 \\
\texttt{LLaMA-3-8B}  \small{\textsc{5-shot Bm25}}        & 18.15 & 8.48 &  & 12.28 & 10.24 & & 8.32 & 16.15 & & 4.47 & 21.11 & & 17.71 & 6.71 \\
\midrule
{} & \multicolumn{14}{c}{\textit{Our Models}} \\
\midrule
\texttt{LLaMA-2-7B} uni.    & 13.00 & 14.76 &  & 13.11 & 8.57 & & 12.30 & 10.97 & & 7.30 & 15.87 & & 15.11 & 7.98 \\
\texttt{LLaMA-2-7B} uni. \small{\textit{beam size=5}}  & 14.93 & 12.55 &  & 13.77 & 8.42 & & 12.90 & 10.04 & & 7.61 & 14.73 & & 16.08 & 6.31 \\
\texttt{LLaMA-3-8B} uni.    & 23.70 & 6.25 &  & 19.65 & 5.20 & & 16.28 & 7.31 & & 11.76 & 9.91 & & 21.88 & 4.21 \\
\texttt{LLaMA-3-8B} uni. \small{\textit{beam size=5}}  & \bf 25.64 & \bf 5.27 &  & \bf 20.52 & \bf 5.07 & & \bf 17.02 & \bf 6.54 & & \bf 13.60 & \bf 8.51 & & \bf 23.24 & \bf 3.77 \\
\texttt{LLaMA-3-8B} multi.   & 21.77 & 6.95 &  & 18.22 & 6.05 & & 15.54 & 7.96 & & 9.76 & 12.96 & & 20.84 & 4.52 \\
\texttt{LLaMA-3-8B} multi. \small{\textit{beam size=5}}    & 24.07 & 5.68 &  & 19.36 & 5.76 & & 16.09 & 7.13 & & 11.62 & 11.40 & & 22.37 & 3.94 \\
\end{tabular}
}
\resizebox{\textwidth}{!}{
\begin{tabular}{lrrrrrrrrrrrrrr}
\toprule
\multirow{2}{*}{Models}  & \multicolumn{2}{c}{Somali} & & \multicolumn{2}{c}{Sundanese} & & \multicolumn{2}{c}{Swahili} & & \multicolumn{2}{c}{Urdu} & & \multicolumn{2}{c}{Xhosa}\\
\cmidrule{2-3} \cmidrule{5-6} \cmidrule{8-9} \cmidrule{11-12} \cmidrule{14-15}
{} & {BLEU} & {MetricX} &  & {BLEU} & {MetricX} & & {BLEU} & {MetricX} & & {BLEU} & {MetricX} & & {BLEU} & {MetricX}\\
\midrule
{} & \multicolumn{14}{c}{\textit{Toplines}} \\
\midrule
\texttt{NLLB-200-3.3B}       & 17.40 & 4.82 &  & 21.44 & 4.97 & & 36.20 & 4.59 & & 28.41 & 4.44 & & 23.24 & 3.80 \\
\texttt{Gemma-3-27B-It}      & 13.58 & 5.50 &  & 17.01 & 4.77 & & 35.16 & 3.80 & & 26.24 & 3.13 & & 12.82 & 7.62 \\
\midrule
{} & \multicolumn{14}{c}{\textit{Baselines}} \\
\midrule
\texttt{Gemma-2-27B-It}      & 8.93 & 10.77 &  & 14.60 & 7.18 & & 35.99 & 3.78 & & 23.28 & 4.01 & & 9.27 & 12.25 \\
\texttt{LLaMA-3.1-70B It}    & 9.74 & 10.52 &  & 17.34 & 5.13 & & 34.26 & 4.46 & & 27.37 & 3.64 & & 6.67 & 16.01 \\
\texttt{Gemma-2-9B-It}       & 6.32 & 14.67 &  & 12.54 & 8.48 & & 29.63 & 5.23 & & 19.26 & 5.25 & & 6.99 & 18.32 \\
\texttt{Command-R7B}         & 1.85 & 20.20 &  & 8.41 & 5.53 & & 6.51 & 18.85 & & 3.75 & 13.14 & & 2.33 & 22.77 \\
\texttt{Aya-expanse-32B}     & 5.88 & 15.13 &  & 9.81 & 10.84 & & 9.37 & 17.25 & & 11.14 & 7.72 & & 4.84 & 22.00 \\
\texttt{Qwen-2.5-32B-It}     & 4.11 & 19.35 &  & 8.12 & 15.01 & & 9.46 & 17.44 & & 11.72 & 9.31 & & 4.13 & 22.32 \\
\midrule
\texttt{LLaMAX3-8B Alpaca}      & 11.12 & 7.41 &  & 11.63 & 8.84 & & 26.76 & 6.63 & & 19.94 & 5.72 & & 11.01 & 10.31 \\
\midrule
\texttt{LLaMA-2-7B}  \small{\textsc{5-shot Bm25}}        & 2.05 & 21.97 &  & 6.43 & 16.65 & & 2.85 & 22.86 & & 2.65 & 19.10 & & 2.34 & 23.42 \\
\texttt{LLaMA-3-8B}  \small{\textsc{5-shot Bm25}}        & 4.74 & 18.45 &  & 14.17 & 8.90 & & 22.61 & 8.63 & & 17.17 & 6.47 & & 3.22 & 22.53 \\
\midrule
{} & \multicolumn{14}{c}{\textit{Our Models}} \\
\midrule
\texttt{LLaMA-2-7B} uni.    & 8.89 & 11.73 &  & 14.70 & 7.19 & & 19.19 & 11.42 & & 14.89 & 8.43 & & 9.25 & 15.24 \\
\texttt{LLaMA-2-7B} uni. \small{\textit{beam size=5}}   & 8.20 & 10.63 &  & 15.42 & 5.75 & & 20.96 & 9.52 & & 16.33 & 6.63 & & 7.67 & 13.09 \\
\texttt{LLaMA-3-8B} uni.    & 13.20 & 7.00 &  & 16.84 & 4.88 & & 30.99 & 5.23 & & 22.43 & 4.16 & & 12.39 & 9.33 \\
\texttt{LLaMA-3-8B} uni.  \small{\textit{beam size=5}}   & \bf 13.70 & \bf 6.17 &  & \bf 18.16 & \bf 4.26 & & \bf 33.49 & \bf 4.51 & & \bf 23.51 & \bf 3.78 & & \bf 13.93 & \bf 7.77 \\
\texttt{LLaMA-3-8B} multi.   & 12.03 & 7.94 &  & 16.24 & 5.66 & & 28.53 & 6.03 & & 21.65 & 4.54 & & 11.74 & 10.56 \\
\texttt{LLaMA-3-8B} multi. \small{\textit{beam size=5}}    & 12.71 & 7.05 &  & 17.32 & 4.83 & & 31.11 & 5.07 & & 22.92 & 3.96 & & 13.15 & 8.61 \\
\bottomrule
\end{tabular}
}
\end{center}
\caption{BLEU and MetricX scores for 10 English $\rightarrow$ X directions from FLORES~200\iffalse~\citep{goyal-etal-2022-flores, nllb2022}\fi. Best results after fine-tuning \iffalse(including any results that are not statistically worse)\fi are highlighted in bold.}
\label{tab:flores200}
\end{table*}

\subsection{Experimental Setup}
We work on translation from English to 10 LRLs: Basque, Hausa, Igbo, Kinyarwanda, Nepali, Somali, Sundanese, Swahili, Urdu and Xhosa.
\paragraph{Datasets.} FLORES-200~\citep{goyal-etal-2022-flores, nllb2022} is a dataset consisting of translations from web articles into 204 languages. These sentences are divided into two splits: devtest and dev. We use the devtest set (1012 examples) as the evaluation set and the dev set (997 examples) as the selection pool for few-shot MT.
\paragraph{Models.} We use \texttt{Gemma-3-27b-It} \citep{gemmateam2025gemma3technicalreport} as the generator, and \texttt{NLLB-200-3.3B} \citep{nllb2022} as the back-translator in order to reduce the computational cost, as translating in a few-shot setting with \texttt{Gemma-3-27b-It} would be significantly more expensive. We use \texttt{LLaMA-2-7B} \citep{touvron2023llama2openfoundation} and \texttt{LLaMA-3-8B} \citep{dubey2024llama3herdmodels} during fine-tuning and compare the resulting models against strong multilingual LLMs including \texttt{LLaMA-3.1-8B-It} \& \texttt{LLaMA-3.1-70B It} \citep{dubey2024llama3herdmodels}, \texttt{Gemma-2-9B-It} \& \texttt{Gemma-2-27B-It} \citep{gemmateam2024gemma2improvingopen}, \texttt{Aya-expanse-8B} \& \texttt{Aya-expanse-32B} \citep{dang2024ayaexpansecombiningresearch}, \texttt{Qwen-2.5-7B-It} \& \texttt{Qwen-2.5-32B-It} \citep{qwen2, qwen2.5} and \texttt{Command-R7B} \citep{cohere2025commandaenterprisereadylarge}.

\paragraph{Evaluation Metrics.}
We mainly evaluate using MetricX-24~\citep{juraska-etal-2024-metricx}. We use the reference-based version \texttt{MetricX-24-XXL} (which supports the same 101 languages as mT5~\citep{xue2021mt5}). MetricX assigns a score ranging from 0 to 25, with higher scores indicating more errors in the translation. We also use $n$-gram matching metrics via sacreBLEU~\citep{post-2018-call}, namely BLEU\footnote{nrefs:1$|$case:mixed$|$eff:no$|$tok:flores200$|$smooth:exp$|$version:2.4.2}~\citep{10.3115/1073083.1073135} and chrF++\footnote{nrefs:1$|$case:mixed$|$eff:yes$|$nc:6$|$nw:2$|$space:no$|$version:2.4.2}~\citep{popovic-2015-chrf, popovic:2017:WMT} for transparency reasons in Appendix~\ref{appendix:chrf_and_xcomet}.

\paragraph{Implementation Details.}
The generator's temperature of generation is set to 1.0. We back-translate with beam search (beam size = 5). For the topics, we use a list of 67,573 Wikipedia topics curated by \citet{mwcat}. We fine-tune unidirectional models for 5k steps (about 3 hours on 1 H100 80G) with a learning rate of 1e-5, a batch size of 4 with 4 gradient accumulation steps and a maximum sequence length equal to 512. 
The multidirectional model requires 100k steps (about 30 hours on 1 H100 80G) and in both cases we choose the last checkpoint as the final model.
For the statistical significant comparison we follow \citep{koehn-2004-statistical} and use paired bootstrap resampling with 300 samples of 500 sentences and a $p$-value threshold of 0.05. All models are evaluated in a zero-shot fashion with greedy decoding unless stated otherwise. See Appendix~\ref{appendix:resources} and \ref{appendix:details} for additional details.

\section{Results}

\subsection{Fine-tuning}

We fine-tune two small models (\texttt{LLaMA-2-7B} and \texttt{LLaMA-3-8B}) on our constructed dataset and compare them to state-of-the-art models of various sizes in zero-shot.\footnote{We use 5-shot with BM25 selection in the FLORES dev set for base models as they cannot follows instructions.} We evaluate two setups: a unidirectional setting with one model per translation direction (English$\leftrightarrow$X), and a multidirectional setting with a single model trained on all 10 directions. Results are shown in Table~\ref{tab:flores200}\footnote{See Appendix~\ref{appendix:to_english} for MT into English.}. Fine-tuning \texttt{LLaMA-2-7B}, whose performance is close to random in all directions turns it into a model that outperforms \texttt{Aya-expanse-32B}, \texttt{Qwen-2.5-32B} and \texttt{Command-R7B}. Unidirectional fine-tuning of \texttt{LLaMA-3-8B} outperforms \texttt{Gemma-2-27B-It} and \texttt{LLaMA-3.1-70B-It}. With beam search \citep{freitag-al-onaizan-2017-beam}, we get even better results with unidirectional models, closing the gap with the generator. Fine-tuning a model to support the ten languages together leads to a drop of performance of about 1 BLEU in all languages, i.e.~there is no positive cross-lingual transfer. We provide more baseline comparisons in Appendix~\ref{appendix:other_baselines}. Additional results from fine-tuning \texttt{NLLB-200-3.3B} and \texttt{Gemma-3-27B-PT} on the \textsc{TopXGen} dataset are provided in Appendix~\ref{appendix:teachers}.

\subsection{In-Context Learning}
We compare the performance obtained when doing 5-shot MT with example selection via similarity search in the FLORES dev set and in the \textsc{TopXGen} dataset with \texttt{LLaMA-3.1-8B-It}. As shown in Table~\ref{tab:flores200_icl}, retrieval in the \textsc{TopXGen} dataset yields superior results compared to zero-shot showing that its content is qualitative enough to help the model during the translation. Moreover, it also works better than retrieval in FLORES, particularly in terms of MetricX. While \textsc{TopXGen} has its size and diversity as advantage, the FLORES dev set is in-domain with respect to the evaluation set as they come from the same research effort on top of being written by professional translators. 

\subsection{Comparison to existing approaches}
\begin{figure}[ht]
\begin{center}
    \includegraphics[width=\linewidth]{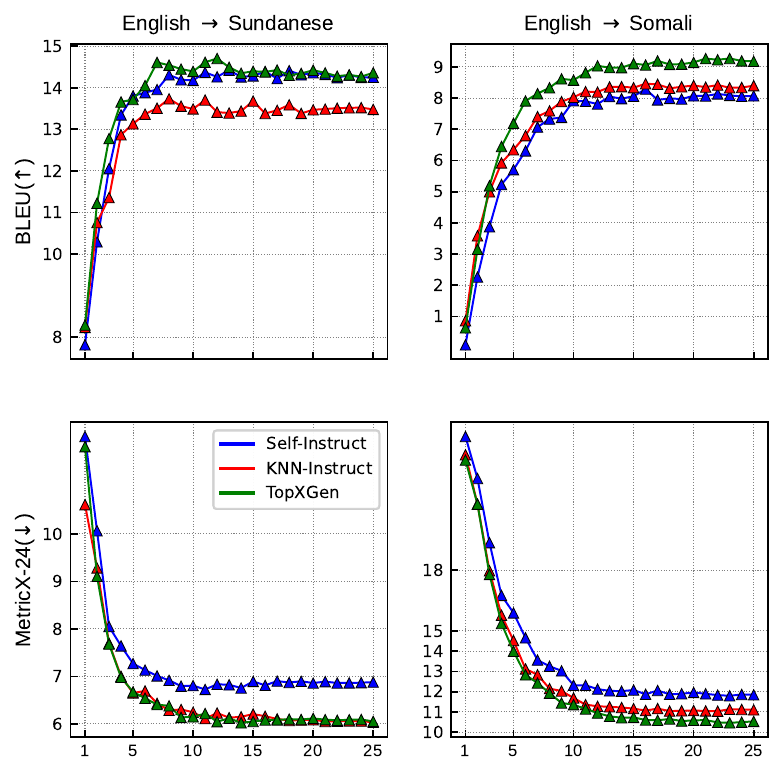}
    \caption{\textsc{TopXGen} vs \textsc{Self-Instruct} \& \textsc{KNN-Instruct}.}
    \label{fig:comparison}
\end{center}
\end{figure}

We investigate the impact of changing the data generation pipeline from our \textsc{TopXGen} to \textsc{Self-Instruct}~\citep{wang-etal-2023-self-instruct} and \textsc{KNN-Instruct}~\citep{kou-etal-2024-knn}. We keep the same parameters (Generator, Seed sentences, Back-translator etc.) and compare the result of fine-tuning \texttt{LLaMA-2-7B} on 20K sentence pairs generated by each strategy in Sundanese and Somali. For \textsc{KNN-Instruct}, we use the multilingual SONAR embeddings \citep{duquenne2023sonarsentencelevelmultimodallanguageagnostic} to compute similarity between sentences and we perform 3 rounds with $K=8$. In Figure~\ref{fig:comparison}, we report the zero-shot (beam size = 5) BLEU and MetricX scores every 200 steps and compare the values across methods. \textsc{TopXGen} consistently outperforms \textsc{Self-Instruct} and \textsc{KNN-Instruct} in terms of BLEU and MetricX at each checkpoint.

\begin{table*}[ht]
\captionsetup{skip=0pt}
\vskip 0.15in
\small
\begin{center}
\resizebox{\textwidth}{!}{
\begin{tabular}{lrrrrrrrrrrrrrr}
\toprule
\multirow{2}{*}{Models}  & \multicolumn{2}{c}{Basque} & & \multicolumn{2}{c}{Hausa} & & \multicolumn{2}{c}{Igbo} & & \multicolumn{2}{c}{Kinyarwanda} & & \multicolumn{2}{c}{Nepali}\\
\cmidrule{2-3} \cmidrule{5-6} \cmidrule{8-9} \cmidrule{11-12} \cmidrule{14-15}
{} & {BLEU} & {MetricX} &  & {BLEU} & {MetricX} & & {BLEU} & {MetricX} & & {BLEU} & {MetricX} & & {BLEU} & {MetricX}\\
\midrule
\texttt{LLaMA-3.1-8B-It} \small{\textsc{Zero-shot}}        & 15.47 & 11.15 &  & 9.61 & 11.75 & & 7.75 & 17.39 & & 3.73 & 20.71 & & 12.21 & 7.66 \\
\texttt{LLaMA-3.1-8B-It}  \small{\textsc{5-shot FLORES}}   & 17.53 & 9.08 &  & 11.85 & 9.98 & & \underline{9.88} & 14.51 & & \bf 5.97 & 19.57 & & \bf 17.18 & 6.37 \\
\texttt{LLaMA-3.1-8B-It}  \small{\textsc{5-shot TopXGen}}  & \bf 18.05 & \bf 8.61 &  & \bf 12.29 & \bf 9.02 & & \bf 9.93 & \bf 13.66 & & 5.69 & \bf 19.16 & & \underline{17.16} & \bf 6.02 \\
\end{tabular}
}
\resizebox{\textwidth}{!}{
\begin{tabular}{lrrrrrrrrrrrrrr}
\toprule
\multirow{2}{*}{Methods}  & \multicolumn{2}{c}{Somali} & & \multicolumn{2}{c}{Sundanese} & & \multicolumn{2}{c}{Swahili} & & \multicolumn{2}{c}{Urdu} & & \multicolumn{2}{c}{Xhosa}\\
\cmidrule{2-3} \cmidrule{5-6} \cmidrule{8-9} \cmidrule{11-12} \cmidrule{14-15}
{} & {BLEU} & {MetricX} &  & {BLEU} & {MetricX} & & {BLEU} & {MetricX} & & {BLEU} & {MetricX} & & {BLEU} & {MetricX}\\
\midrule
\texttt{LLaMA-3.1-8B-It} \small{\textsc{Zero-shot}}       & 4.20 & 19.63 &  & 10.30 & 10.71 & & 20.44 & 9.21 & & 19.24 & 5.93 & & 3.41 & 23.41 \\
\texttt{LLaMA-3.1-8B-It} \small{\textsc{5-shot FLORES}}   & 6.04 & 17.12 &  & \bf 14.06 & 8.56 & & 23.47 & 8.69 & & 20.75 & 5.54 & & \bf 4.67 & 22.00 \\
\texttt{LLaMA-3.1-8B-It} \small{\textsc{5-shot TopXGen}}  & \bf 6.50 & \bf 15.39 &  & 13.28 & \bf 8.14 & & \bf 24.14 & \bf 8.22 & & \bf 21.06 & \bf 5.26 & & \underline{4.63} & \bf 21.08 \\
\bottomrule
\end{tabular}
}
\end{center}
\caption{BLEU and MetricX scores for 10 English $\rightarrow$ X directions from FLORES~200 with ICL. Best results are highlighted in bold. Scores that are statistically equivalent to the best are underlined.}
\label{tab:flores200_icl}
\end{table*}

\begin{figure}[ht]
\begin{center}
    \includegraphics[width=\linewidth]{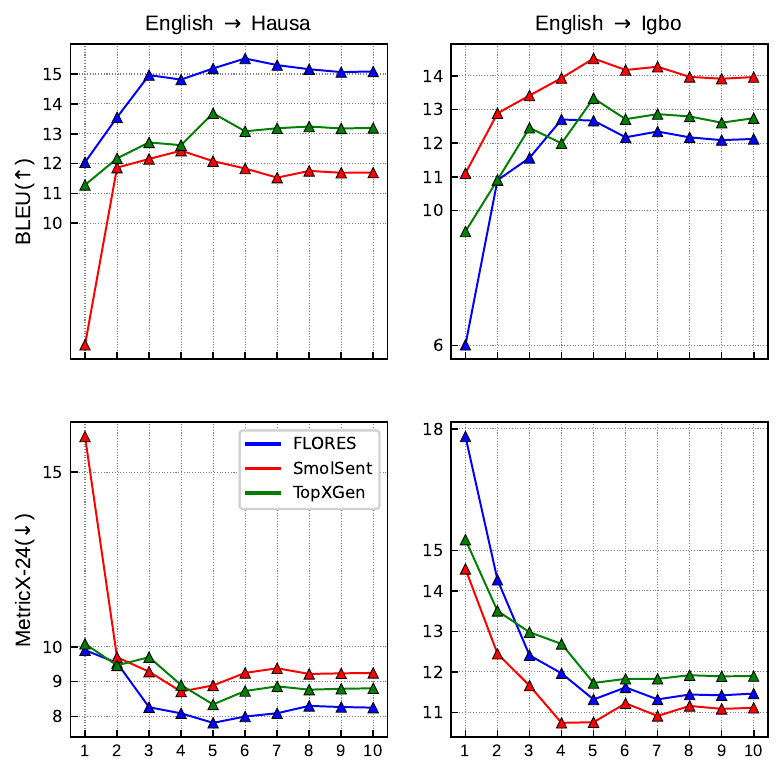}
    \caption{\textsc{TopXGen} vs FLORES \& \textsc{SmolSent}.}
    \label{fig:real}
\end{center}
\end{figure}

We also compare fine-tuning on \textsc{TopXGen} to fine-tuning on high-quality professionally translated data such as \textsc{SmolSent} \citep{jones2023gatitos, caswell2025smol} (863 parallel sentences) and the FLORES dev set (997 samples). We select the first 863 samples from each dataset and choose \texttt{LLaMA-3-8B} as the base model due to the limited data. We fine-tune one model for English $\leftrightarrow$ Hausa and another for English $\leftrightarrow$ Igbo and report the scores every 100 steps on Figure~\ref{fig:real}. As expected, \textsc{TopXGen} does not consistently work better than professional translations. However it outperforms \textsc{SmolSent} when translating in Hausa and competes with FLORES in Igbo. Moreover, \textsc{TopXGen} has the advantage of the scale, as proven in Table~\ref{tab:flores200}, fine-tuning on full \textsc{TopXGen} dataset (which is 100 times bigger) outperforms the best \textsc{SmolSent} and FLORES checkpoint by at least 3 BLEU.

\section{Ablation Studies}

\paragraph{Impact of the generator and the number of topics.}
In this section, we focus on Sundanese and study 2 setups. First, we use \texttt{gpt-4o-mini-2024-07-18}~\citep{OpenAI_GPT-4} as the generator LLM and analyze the performance of small models (\texttt{LLaMA-2-7B} and \texttt{LLaMA-3-8B}) fine-tuned on 55k sentences. Second, we reduce the number of topics from 67,573 to a curated list of 509 elements and analyze how the fine-tuned models perform for both generator(s). As shown in Table~\ref{tab:teacher}, a stronger generator (here \texttt{gpt-4o-mini}) results in stronger students. Moreover, we observe that fine-tuning with more topics achieves stronger results, suggesting that diversity matters for building strong student models. The number of topics is also an important aspect when you generate data on a larger scale, beyond the numbers that we explore in this paper.

\begin{table}[ht]
\small
\begin{center}
\resizebox{\linewidth}{!}{
\begin{tabular}{lrrrrrrrrrrrrrr}
\toprule
\multirow{2}{*}{Models}  & \multicolumn{2}{c}{Less Topics} & & \multicolumn{2}{c}{Usual Topics} \\
\cmidrule{2-3} \cmidrule{5-6} \cmidrule{8-9}
{} & {BLEU} & {MetricX} &  & {BLEU} & {MetricX} \\
\midrule
\texttt{Gemma 3 27B It}          & 17.01 & 4.77 &  & 17.01  & 4.77 \\
\midrule
\texttt{LLaMA-2-7B} uni. Gemma  & 14.03 & 6.04 &  & 15.31 & 5.72 \\
\texttt{LLaMA-3-8B} uni. Gemma  & 16.97 & 4.46 &  & \bf 18.15 & \bf 4.35 \\
\midrule
\texttt{gpt-4o-mini-2024-07-18}  & 24.17 & 3.46 &  & 24.17 & 3.46 \\
\midrule
\texttt{LLaMA-2-7B} uni. GPT    & 17.65 & 5.79 &  & 18.25 & 5.65 \\
\texttt{LLaMA-3-8B} uni. GPT    & 19.79 & 4.21 &  & \bf 20.51 & \bf 4.09 \\
\bottomrule
\end{tabular}
}
\end{center}
\caption{Results for English $\rightarrow$ Sundanese direction on FLORES~200 (BLEU and MetricX scores) for different generators and topics.}
\label{tab:teacher}
\end{table}

\paragraph{Impact of the Back-translator.}
Throughout the paper, we used \texttt{NLLB-200-3.3B} as the back-translator. In this section, we study two setups. First, we analyze how the results change when the generator is also used as the back-translator. Back-translation is performed with greedy 5-shot with BM25 retrieval~\citep{robertson1995okapi, bm25s} from the FLORES-200 dev set following \citep{zebaze-etal-2025-context}. We fine-tune separate models on Hausa and Sundanese, and compare their results with using \texttt{NLLB-200-3.3B} as the back-translator in Table~\ref{tab:bt}. Using the generator as the back-translator works almost as well as using \texttt{NLLB-200-3.3B}. This is a direct by-product of the fluency and translation literacy of state-of-the-art LLMs in English. 

\begin{table}[ht]
\captionsetup{skip=0pt}
\small
\begin{center}
\resizebox{\linewidth}{!}{
\begin{tabular}{lrrrrrrrrrrrrrr}
\toprule
\multirow{2}{*}{Methods}  & \multicolumn{2}{c}{Hausa} & & \multicolumn{2}{c}{Sundanese} \\
\cmidrule{2-3} \cmidrule{5-6} \cmidrule{8-9}
{} & {BLEU} & {MetricX} &  & {BLEU} & {MetricX} \\
\midrule
\texttt{LLaMA-2-7B} uni. \textsc{nllb}  & \bf 13.77 & \bf 8.42 &  & \bf 15.42 & \bf 5.75 \\
\texttt{LLaMA-2-7B} uni. \textsc{gemma} & 13.14 & 9.31 &  & 15.19 & \underline{5.83} \\
\midrule
\texttt{LLaMA-3-8B} uni. \textsc{nllb}  & \bf 20.52 & \bf 5.07 &  & \bf 18.16 & \bf 4.26 \\
\texttt{LLaMA-3-8B} uni. \textsc{gemma} & 19.58 & 5.64 &  & 17.98 & \underline{4.31} \\
\bottomrule
\end{tabular}
}
\end{center}
\caption{Full quantitative results for English $\leftrightarrow$ Hausa and English $\leftrightarrow$ Sundanese on FLORES~200 (BLEU and MetricX scores). Best results are highlighted in bold. Scores that are statistically equivalent to the best are underlined.}
\label{tab:bt}
\end{table}
Second, we use the [fine-tuned] student as the back-translator (also in 5-shot) with the idea of performing iterative self-improvement~\citep{NEURIPS2022_639a9a17, pmlr-v235-chen24j} via iterative back-translation~\citep{NIPS2016_5b69b9cb, lample-etal-2018-phrase, artetxe-etal-2018-unsupervised}. For a language $l$, we start by using $M_0 = \texttt{LLaMA-2-7B}$ to back-translate $Y^l$ (generated via \textsc{TopXGen}) and create $X_0$. We create $M_1$ by fine-tuning $M_0$ on $(Y^l \rightarrow X_0) \cup (X_0 \rightarrow Y^l)$. Similarly, we create $X_1$ by back-translating $Y^l$ with $M_1$, build $M_2$ by fine-tuning $M_0$ on $(Y^l \rightarrow X_1) \cup (X_1 \rightarrow Y^l)$ and so on. The same model performs the translation into both directions. We apply that pipeline with \texttt{LLaMA-2-7B} and \texttt{LLaMA-3-8B} on data generated by \texttt{gpt-4o-mini} in Sundanese (Table~\ref{tab:iterative}). At round 0—when the base student model also serves as the back-translator—we already observe improvements in both translation directions, though they are smaller than when using NLLB as the back-translator. After one iteration, \texttt{LLaMA-2-7B} shows further gains, with a +3 BLEU increase and a -1 MetricX drop for English-to-Sundanese, and a modest improvement in the reverse direction. However, the self-improvement process soon plateaus, largely because performance on the English side stagnates.
In contrast, \texttt{LLaMA-3-8B} fails to significantly improve Sundanese-to-English performance at round 0, preventing the iterative process from taking off. We hypothesize that self-improvement stalls when the model's English-side performance nears that of the generator on the Sundanese side. Thus, initializing the pipeline with higher-quality Sundanese data than that provided by \textsc{TopXGen} could potentially support more improvement and additional self-iteration rounds.

\begin{table}[ht]
\captionsetup{skip=0pt}
\small
\begin{center}
\resizebox{\linewidth}{!}{
\begin{tabular}{lrrrrrrrr}
\toprule
\multirow{2}{*}{Models}  & \multicolumn{2}{c}{English to Sundanese} & & \multicolumn{2}{c}{Sundanese to English} \\
\cmidrule{2-3} \cmidrule{5-6} \cmidrule{8-9}
{} & {BLEU} & {MetricX} &  & {BLEU} & {MetricX} \\
\midrule
\texttt{gpt-4o-mini}        & 24.17 & 3.46 &  & 39.66 & 2.67 \\
\midrule
\texttt{LLaMA-2-7B}                              & 6.43  & 16.65 &  & 16.30 & 8.76 \\
Round 0                                          & 10.78 & 6.86 &  & 21.84 & 6.41 \\
Round 1                                          & 14.08 & 5.82 &  & \bf 22.36 & \bf 6.37 \\
Round 2                                          & \bf 14.30 & 5.79 &  & 22.25 & 6.53 \\
Round 3                                          & 14.10 & \bf 5.69 &  & 21.85 & 6.52 \\
\midrule
\texttt{LLaMA-3-8B}                               & 14.17 & 8.90 &  & 33.77 & 3.83 \\
Round 0                                          & 19.71 & \bf 3.92 &  & \bf 35.55 & \underline{3.48} \\
Round 1                                          & 19.70 & 4.00 &  & \underline{35.37} & \underline{3.48} \\
Round 2                                          & 19.75 & \underline{3.94} &  & \underline{35.40} & \bf 3.46 \\
Round 3                                          & \bf 19.84 & \underline{3.95} &  & \underline{35.44} &  \underline{3.47} \\
\bottomrule
\end{tabular}
}
\end{center}
\caption{Results for English $\leftrightarrow$ Sundanese direction on FLORES-200 (BLEU and MetricX scores) in 5-shot during Self-Improvement. Best results are highlighted in bold. Scores that are statistically equivalent to the best are underlined.}
\label{tab:iterative}
\end{table}

\begin{figure}[ht]
\begin{center}
    \includegraphics[width=\linewidth]{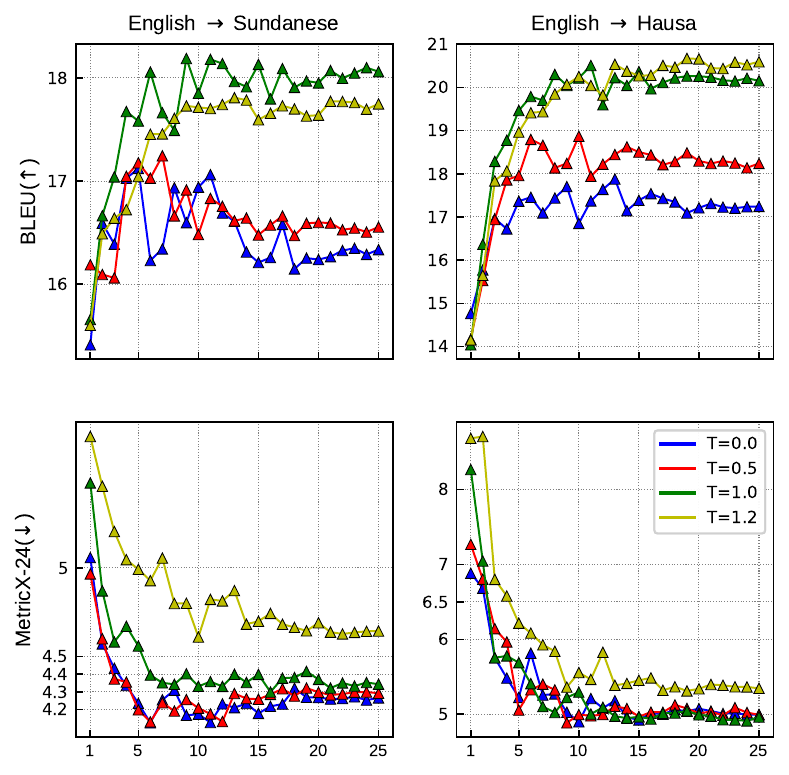}
    \caption{Impact of the temperature of the Generator.}
    \label{fig:temperature}
\end{center}
\end{figure}

\paragraph{Impact of the temperature of the generation of the generator.} A high temperature generally correlates with more diversity in the generations, but working with LRLs, there is a risk of going off the track in terms of quality. Here we consider four temperatures and analyze the zero-shot (beam size = 5) performance on FLORES-200 every 200 steps after respectively fine-tuning \texttt{LLaMA-3-8B} on 50K and 40K sentence pairs generated by \textsc{TopXGen} with \texttt{Gemma-3-27B-It} respectively in Sundanese and Hausa. As illustrated in Figure~\ref{fig:temperature}, generating with a temperature of $T=1.0$ leads to a stronger model after fine-tuning, yielding up to 3 BLEU points more than $T=0.0$ and $T=0.5$, while maintaining comparable MetricX scores. Using a temperature of $T = 1.2$ does not yield improvements over $T = 1.0$ in terms of BLEU, and results in lower MetricX scores compared to all other temperature values. Our attempts to generate content with a temperature greater than 1.2 (typically 1.5, 2.0 and 5.0) resulted into garbage generations: nonsensical or invented ``sentences'' that mix morphemes, word fragments, and orthographic features from multiple scripts and languages.

\section{Conclusion}
We presented \textsc{TopXGen}, a scalable pipeline for constructing synthetic parallel datasets for MT into LRLs. 
Our method generates diverse monolingual data directly in a wide range of LRLs using topic-guided prompting \citep{he2024exploring, aycock-bawden-2024-topic}, which we then back-translate into a HRL, building synthetic parallel datasets. Our experiments demonstrate that models trained on data generated via \textsc{TopXGen} achieve strong performance across both unidirectional and multidirectional scenarios, approaching the generator's performance while requiring only small data volumes and limited GPU resources. Furthermore, existing generation strategies such as \textsc{Self-Instruct} and \textsc{KNN-Instruct} are compatible with our pipeline but fall short in terms of performance. \textsc{TopXGen} enables the creation of synthetic datasets which are comparable to professionally written text for ICL and fine-tuning, offering a practical and effective solution for low-resource multilingual MT.

\section*{Limitations}
In this paper, we demonstrate that it is possible to use synthetic data to get smaller language models to improve their translation capabilities into low-resource languages. However, one limitation is the requirement of a language model (generator) which ``understands pretty well'' the languages of interest. As suggested in the paper, using monolingual data scrapped from the internet can alleviate such a limitation but it can be hard to find or manually build high-quality monolingual data covering a wide variety of topics in some languages.

\section*{Acknowledgments}
This work was partly funded by Rachel Bawden and Benoît Sagot's chairs in the PRAIRIE institute, now PRAIRIE-PSAI, funded by the French national agency ANR, respectively as part of the “Investissements d’avenir” programme under the reference ANR-19-P3IA-0001 and as part of the ``France 2030'' strategy under the reference ANR-23-IACL-0008.
This work was granted access to the HPC resources of IDRIS under the allocation 2025-AD011015933 made by GENCI.


\bibliography{custom}

\appendix

\section{Reproducibility Details} \label{appendix:reproducibility_details}

\subsection{Models, Datasets and Tools} \label{appendix:resources}

In Table~\ref{tab:url}, we list the links to the relevant resources used for the experiments.
\begin{table*}[!ht]
    \centering\small
    \resizebox{\linewidth}{!}{
    \begin{tabular}{l l}
    \toprule
    \multicolumn{2}{c}{\textit{Datasets}} \\
    \midrule
    FLORES-200 & \url{https://huggingface.co/datasets/facebook/flores} \\
    NTREX HF & \url{hhttps://huggingface.co/datasets/mteb/NTREX} \\
    TICO-19 & \url{https://huggingface.co/datasets/gmnlp/tico19}\\
    \midrule
    \multicolumn{2}{c}{\textit{Models evaluated}} \\
    \midrule
    \texttt{Aya-expanse-8B}  & \url{https://huggingface.co/CohereLabs/aya-expanse-8b} \\
    \texttt{Aya-expanse-32B} & \url{https://huggingface.co/CohereLabs/aya-expanse-32b} \\
    \texttt{Command-R7B}     & \url{https://huggingface.co/CohereLabs/c4ai-command-r7b-12-2024}\\
    \texttt{Qwen2.5-7B-Instruct} & \url{https://huggingface.co/Qwen/Qwen2.5-7B-Instruct} \\
    \texttt{Qwen2.5-32B-Instruct} & \url{https://huggingface.co/Qwen/Qwen2.5-32B-Instruct} \\
    \texttt{LLaMAX2-7B-Alpaca} & \url{https://huggingface.co/LLaMAX/LLaMAX2-7B-Alpaca} \\
    \texttt{LLaMAX3-8B-Alpaca} & \url{https://huggingface.co/LLaMAX/LLaMAX3-8B-Alpaca} \\
    \texttt{LLaMA-2-7B} & \url{https://huggingface.co/meta-llama/Llama-2-7b-hf} \\
    \texttt{LLaMA-3-8B} & \url{https://huggingface.co/meta-llama/Meta-Llama-3-8B} \\
    \texttt{Gemma-2-9B-It} & \url{https://huggingface.co/google/gemma-2-9b-it} \\
    \texttt{Gemma-2-27B-It} & \url{https://huggingface.co/google/gemma-2-27b-it} \\
    \texttt{Gemma-3-4B-It} & \url{https://huggingface.co/google/gemma-3-4b-it} \\
    \texttt{Gemma-3-27B-It} & \url{https://huggingface.co/google/gemma-3-27b-it} \\
    \texttt{Gemma-3-4B-Pt} & \url{https://huggingface.co/google/gemma-3-4b-pt} \\
    \texttt{Gemma-3-27B-Pt} & \url{https://huggingface.co/google/gemma-3-27b-pt} \\
    \texttt{LLaMA-3.1-8B-It} & \url{https://huggingface.co/meta-llama/Meta-Llama-3.1-8B-Instruct} \\
    \texttt{LLaMA-3.1-70B-It} & \url{https://huggingface.co/meta-llama/Llama-3.1-70B-Instruct}\\
    \texttt{NLLB-200-3.3B} & \url{https://huggingface.co/facebook/nllb-200-3.3B}\\
    \midrule
    \multicolumn{2}{c}{\textit{Other resources}} \\
    \midrule
    MetricX24-XXL & \url{https://huggingface.co/google/metricx-24-hybrid-xxl-v2p6} \\
    XCOMET-XL & \url{https://huggingface.co/Unbabel/XCOMET-XL} \\
    BM25s & \url{https://github.com/xhluca/bm25s} \\
    FastText & \url{https://huggingface.co/facebook/fasttext-language-identification} \\
    Wikipedia topics & \url{https://huggingface.co/datasets/tarekziade/wikipedia-topics} \\
    \texttt{vLLM} \citep{kwon2023efficient} & \url{https://github.com/vllm-project/vllm} \\
    \bottomrule
    \end{tabular}
    }
    \caption{Links to datasets, benchmarks and models.}
    \label{tab:url}
\end{table*}

\subsection{Implementation Details} \label{appendix:details}
We use HuggingFace's Transformers library \citep{wolf2020huggingfacestransformersstateoftheartnatural}. For training the unidirectional models, we use a per-device batch size of 4 and apply gradient accumulation over 4 steps. We use stacking with a maximum length of 512 tokens. Training is conducted over 5,000 steps using a learning rate of 1e-5, with 500 warmup steps. The learning rate decays to zero following a cosine schedule. We also apply a weight decay of 0.01. For the multidirectional models, which are trained across the 10 target languages, we use the same hyperparameters but extend training to 100,000 steps. We adopt the prompt template introduced by \citet{xu2024a}, and compute the loss only on the target sentence. In all experiments, we fine-tune in both directions (i.e., source-to-target and target-to-source), so each parallel sentence pair contributes two samples to the training set.
On smaller datasets such as FLORES and \textsc{SmolSent}, we reduce the training steps to 1,000. In this setting, the batch size per device is 2, we use 100 warmup steps and we do not perform gradient accumulation. All other hyperparameters remain unchanged.

\begin{lstlisting}[frame=single,breaklines=true,basicstyle=\small\ttfamily]
Translate this from English to Hausa:
English: "We now have 4-month-old mice that are non-diabetic that used to be diabetic," he added.
Hausa:
\end{lstlisting}

\section{Additional Experiments} \label{appendix:additional_experiments}

\subsection{More baselines} \label{appendix:other_baselines}
\begin{table*}[ht]
\captionsetup{skip=0pt}
\vskip 0.15in
\small
\begin{center}
\resizebox{\textwidth}{!}{
\begin{tabular}{lrrrrrrrrrrrrrr}
\toprule
\multirow{2}{*}{Models}  & \multicolumn{2}{c}{Basque} & & \multicolumn{2}{c}{Hausa} & & \multicolumn{2}{c}{Igbo} & & \multicolumn{2}{c}{Kinyarwanda} & & \multicolumn{2}{c}{Nepali}\\
\cmidrule{2-3} \cmidrule{5-6} \cmidrule{8-9} \cmidrule{11-12} \cmidrule{14-15}
{} & {BLEU} & {MetricX} &  & {BLEU} & {MetricX} & & {BLEU} & {MetricX} & & {BLEU} & {MetricX} & & {BLEU} & {MetricX}\\
\midrule
{} & \multicolumn{14}{c}{\textit{Other Baselines}} \\
\midrule
\texttt{LLaMA-3.3-70B It}    & 25.02 & 5.56 &  & 18.15 & 6.09 & & 14.36 & 9.05 & & 8.11 & 14.05 & & 23.28 & 4.60 \\
\texttt{LLaMA-3.1-8B-It}     & 15.47 & 11.15 &  & 9.61 & 11.75 & & 7.75 & 17.39 & & 3.73 & 20.71 & & 12.21 & 7.66 \\
\texttt{Aya-expanse-8B}      & 4.53 & 19.98 &  & 3.33 & 16.53 & & 3.66 & 22.41 & & 2.25 & 23.11 & & 5.28 & 6.64 \\
\texttt{Qwen-2.5-7B-It}      & 4.31 & 22.04 &  & 4.39 & 19.37 & & 4.13 & 22.72 & & 1.94 & 24.28 & & 5.22 & 12.50 \\
\texttt{LLaMAX2-7B Alpaca}      & 10.56 & 15.38 &  & 17.17 & 6.75 & & 9.34 & 14.59 & & 4.57 & 18.85 & & 12.07 & 14.97 \\
\midrule
{} & \multicolumn{14}{c}{\textit{Our Models}} \\
\midrule
\texttt{LLaMA-2-7B} uni.    & 13.00 & 14.76 &  & 13.11 & 8.57 & & 12.30 & 10.97 & & 7.30 & 15.87 & & 15.11 & 7.98 \\
\texttt{LLaMA-2-7B} uni. \small{\textit{beam size=5}}  & 14.93 & 12.55 &  & 13.77 & 8.42 & & 12.90 & 10.04 & & 7.61 & 14.73 & & 16.08 & 6.31 \\
\texttt{LLaMA-3-8B} uni.    & 23.70 & 6.25 &  & 19.65 & 5.20 & & 16.28 & 7.31 & & 11.76 & 9.91 & & 21.88 & 4.21 \\
\texttt{LLaMA-3-8B} uni. \small{\textit{beam size=5}}  & \bf 25.64 & \bf 5.27 &  & \bf 20.52 & \bf 5.07 & & \bf 17.02 & \bf 6.54 & & \bf 13.60 & \bf 8.51 & & \bf 23.24 & \bf 3.77 \\
\texttt{LLaMA-3-8B} multi.   & 21.77 & 6.95 &  & 18.22 & 6.05 & & 15.54 & 7.96 & & 9.76 & 12.96 & & 20.84 & 4.52 \\
\texttt{LLaMA-3-8B} multi. \small{\textit{beam size=5}}    & 24.07 & 5.68 &  & 19.36 & 5.76 & & 16.09 & 7.13 & & 11.62 & 11.40 & & 22.37 & 3.94 \\
\end{tabular}
}
\resizebox{\textwidth}{!}{
\begin{tabular}{lrrrrrrrrrrrrrr}
\toprule
\multirow{2}{*}{Models}  & \multicolumn{2}{c}{Somali} & & \multicolumn{2}{c}{Sundanese} & & \multicolumn{2}{c}{Swahili} & & \multicolumn{2}{c}{Urdu} & & \multicolumn{2}{c}{Xhosa}\\
\cmidrule{2-3} \cmidrule{5-6} \cmidrule{8-9} \cmidrule{11-12} \cmidrule{14-15}
{} & {BLEU} & {MetricX} &  & {BLEU} & {MetricX} & & {BLEU} & {MetricX} & & {BLEU} & {MetricX} & & {BLEU} & {MetricX}\\
\midrule
{} & \multicolumn{14}{c}{\textit{Other Baselines}} \\
\midrule
\texttt{LLaMA-3.3-70B It}    & 9.35 & 11.14 &  & 17.27 & 5.46 & & 33.89 & 4.75 & & 26.31 & 3.88 & & 6.21 & 16.14 \\
\texttt{LLaMA-3.1-8B-It}     & 4.20 & 19.63 &  & 10.30 & 10.71 & & 20.44 & 9.21 & & 19.24 & 5.93 & & 3.41 & 23.41 \\
\texttt{Aya-expanse-8B}      & 4.46 & 18.08 &  & 9.66 & 3.42 & & 4.82 & 21.81 & & 6.60 & 9.03 & & 3.60 & 23.99 \\
\texttt{Qwen-2.5-7B-It}      & 3.50 & 21.19 &  & 5.90 & 16.69 & & 4.94 & 21.92 & & 5.10 & 13.56 & & 2.55 & 24.28 \\
\texttt{LLaMAX2-7B Alpaca}      & 9.00 & 10.72 &  & 8.60 & 12.13 & & 21.62 & 7.34 & & 9.17 & 17.59 & & 12.03 & 10.31 \\
\midrule
{} & \multicolumn{14}{c}{\textit{Our Models}} \\
\midrule
\texttt{LLaMA-2-7B} uni.    & 8.89 & 11.73 &  & 14.70 & 7.19 & & 19.19 & 11.42 & & 14.89 & 8.43 & & 9.25 & 15.24 \\
\texttt{LLaMA-2-7B} uni. \small{\textit{beam size=5}}   & 8.20 & 10.63 &  & 15.42 & 5.75 & & 20.96 & 9.52 & & 16.33 & 6.63 & & 7.67 & 13.09 \\
\texttt{LLaMA-3-8B} uni.    & 13.20 & 7.00 &  & 16.84 & 4.88 & & 30.99 & 5.23 & & 22.43 & 4.16 & & 12.39 & 9.33 \\
\texttt{LLaMA-3-8B} uni.  \small{\textit{beam size=5}}   & \bf 13.70 & \bf 6.17 &  & \bf 18.16 & \bf 4.26 & & \bf 33.49 & \bf 4.51 & & \bf 23.51 & \bf 3.78 & & \bf 13.93 & \bf 7.77 \\
\texttt{LLaMA-3-8B} multi.   & 12.03 & 7.94 &  & 16.24 & 5.66 & & 28.53 & 6.03 & & 21.65 & 4.54 & & 11.74 & 10.56 \\
\texttt{LLaMA-3-8B} multi. \small{\textit{beam size=5}}    & 12.71 & 7.05 &  & 17.32 & 4.83 & & 31.11 & 5.07 & & 22.92 & 3.96 & & 13.15 & 8.61 \\
\bottomrule
\end{tabular}
}
\end{center}
\caption{BLEU and MetricX scores for 10 English $\rightarrow$ X directions from FLORES~200\iffalse~\citep{goyal-etal-2022-flores, nllb2022}\fi. Best results after fine-tuning \iffalse(including any results that are not statistically worse)\fi are highlighted in bold.}
\label{tab:flores200_more}
\end{table*}
Table~\ref{tab:flores200_more} presents a comparison of our fine-tuned student models on FLORES-200 against additional baselines. They surpass all evaluated 7B models, and notably, unidirectional models based on \texttt{LLaMA-3-8B} with beam search outperform \texttt{LLaMA-3.3-70B It}.

\subsection{Translation into English} \label{appendix:to_english}
In Table~\ref{tab:flores200_reverse}, we present the performance of the fine-tuned student models on FLORES-200 for translation from low-resource languages into English. The results mirror our observations in the reverse direction: unidirectional fine-tuning consistently outperforms multidirectional fine-tuning, and using a strong base model (\texttt{LLaMA-3-8B}) leads to significantly better performance—comparable to models that are up to three times larger.

\begin{table*}[ht]
\vskip 0.15in
\small
\begin{center}
\resizebox{\textwidth}{!}{
\begin{tabular}{lrrrrrrrrrrrrrr}
\toprule
\multirow{2}{*}{Methods}  & \multicolumn{2}{c}{Basque} & & \multicolumn{2}{c}{Hausa} & & \multicolumn{2}{c}{Igbo} & & \multicolumn{2}{c}{Kinyarwanda} & & \multicolumn{2}{c}{Nepali}\\
\cmidrule{2-3} \cmidrule{5-6} \cmidrule{8-9} \cmidrule{11-12} \cmidrule{14-15}
{} & {BLEU} & {MetricX} &  & {BLEU} & {MetricX} & & {BLEU} & {MetricX} & & {BLEU} & {MetricX} & & {BLEU} & {MetricX}\\
\midrule
\texttt{NLLB-200-3.3B}       & 36.21 & 2.46 &  & 36.98 & 4.27 & & 32.88 & 4.92 & & 35.17 & 3.96 & & 44.98 & 2.93 \\
\texttt{Gemma-3-27B-It}      & 35.59 & 1.98 &  & 31.20 & 4.28 & & 25.30 & 5.74 & & 28.59 & 4.45 & & 40.73 & 2.19 \\
\midrule
\texttt{Gemma-2-27B-It}      & 35.07 & 2.18 &  & 28.76 & 4.83 & & 21.23 & 7.14 & & 21.86 & 6.67 & & 39.14 & 2.63 \\
\texttt{LLaMA-3.1-70B It}    & 36.53 & 2.24 &  & 31.80 & 4.75 & & 26.39 & 6.25 & & 23.19 & 7.09 & & 40.74 & 2.84 \\
\midrule
\texttt{Gemma-2-9B-It}       & 32.99 & 2.82 &  & 27.04 & 5.65 & & 18.59 & 8.50 & & 19.45 & 8.28 & & 37.09 & 2.92 \\
\texttt{LLaMA-3.1-8B-It}     & 26.78 & 4.12 &  & 19.91 & 8.69 & & 15.27 & 10.89 & & 12.54 & 12.17 & & 28.03 & 4.97 \\
\texttt{Command-R7B}         & 12.21 & 11.63 &  & 5.72 & 16.84 & & 5.94 & 18.60 & & 6.46 & 17.84 & & 23.25 & 6.49 \\
\texttt{Aya-expanse-32B}     & 24.41 & 5.16 &  & 9.56 & 15.39 & & 8.49 & 15.20 & & 10.79 & 13.99 & & 28.33 & 4.49 \\
\texttt{Aya-expanse-8B}      & 12.13 & 10.34 &  & 6.68 & 16.88 & & 6.37 & 16.23 & & 6.84 & 17.29 & & 18.22 & 6.75 \\
\texttt{Qwen-2.5-32B-It}     & 23.55 & 6.48 &  & 12.10 & 14.63 & & 10.15 & 14.38 & & 10.09 & 15.74 & & 33.27 & 4.37 \\
\texttt{Qwen-2.5-7B-It}      & 13.27 & 11.06 &  & 8.09 & 17.37 & & 7.31 & 17.73 & & 7.53 & 18.99 & & 21.85 & 7.06 \\
\midrule
\texttt{LLaMAX2-7B Alpaca}   & 15.36 & 8.73 &  & 20.16 & 6.77 & & 12.15 & 12.39 & & 10.77 & 11.19 & & 15.28 & 12.19 \\
\texttt{LLaMAX3-8B Alpaca}   & 29.27 & 3.09 &  & 27.84 & 4.56 & & 21.72 & 6.60 & & 17.58 & 7.78 & & 34.02 & 3.21 \\
\midrule
\texttt{LLaMA-2-7B}  \small{\textsc{5-shot Bm25}}        & 8.22 & 13.54 &  & 6.68 & 17.13 & & 5.83 & 17.39 & & 5.98 & 17.53 & & 10.32 & 11.24 \\
\texttt{LLaMA-3-8B}   \small{\textsc{5-shot Bm25}}       & 31.82 & 2.74 &  & 26.93 & 5.61 & & 22.69 & 7.43 & & 16.71 & 9.06 & & 33.43 & 3.76 \\
\midrule
\texttt{LLaMA-2-7B} uni.    & 21.00 & 8.41 &  & 20.82 & 9.79 & & 16.27 & 12.21 & & 16.80 & 11.20 & & 24.29 & 7.43 \\
\texttt{LLaMA-2-7B} uni.  \small{\textit{beam size=5}}    & 21.78 & 7.75 &  & 20.81 & 9.26 & & 15.38 & 11.59 & & 17.09 & 10.77 & & 26.22 & 6.75 \\
\texttt{LLaMA-3-8B} uni.    & 34.71 & 3.21 &  & 32.42 & 6.07 & & 26.22 & 7.84 & & 27.42 & 6.70 & & 38.69 & 3.79 \\
\texttt{LLaMA-3-8B} uni.  \small{\textit{beam size=5}}  & \bf 35.43 & \bf 3.07 &  & \bf 33.07 & \bf 5.81 & & \bf 27.70 & \bf 7.35 & & \bf 28.35 & \bf 6.33 & & \bf 40.00 & \bf 3.62 \\
\texttt{LLaMA-3-8B} multi.   & 34.51 & 3.19 &  & 31.75 & 6.33 & & 26.27 & 8.04 & & 25.38 & 8.17 & & 38.81 & 3.98 \\
\texttt{LLaMA-3-8B} multi. \small{\textit{beam size=5}}  & 35.07 & 3.05 &  & 32.67 & 6.04 & & 27.56 & 7.60 & & 26.34 & 7.74 & & 39.41 & 3.76 \\
\end{tabular}
}
\resizebox{\textwidth}{!}{
\begin{tabular}{lrrrrrrrrrrrrrr}
\toprule
\multirow{2}{*}{Methods}  & \multicolumn{2}{c}{Somali} & & \multicolumn{2}{c}{Sundanese} & & \multicolumn{2}{c}{Swahili} & & \multicolumn{2}{c}{Urdu} & & \multicolumn{2}{c}{Xhosa}\\
\cmidrule{2-3} \cmidrule{5-6} \cmidrule{8-9} \cmidrule{11-12} \cmidrule{14-15}
{} & {BLEU} & {MetricX} &  & {BLEU} & {MetricX} & & {BLEU} & {MetricX} & & {BLEU} & {MetricX} & & {BLEU} & {MetricX}\\
\midrule
\texttt{NLLB-200-3.3B}       & 31.02 & 5.50 &  & 42.04 & 3.33 & & 45.70 & 2.88 & & 39.73 & 3.08 & & 40.06 & 3.71 \\
\texttt{Gemma-3-27B-It}      & 28.76 & 4.57 &  & 37.78 & 2.92 & & 46.09 & 2.19 & & 36.90 & 2.14 & & 31.96 & 5.14 \\
\midrule
\texttt{Gemma-2-27B-It}      & 22.13 & 6.85 &  & 32.60 & 3.74 & & 46.23 & 2.26 & & 37.04 & 2.54 & & 28.17 & 5.76 \\
\texttt{LLaMA-3.1-70B It}    & 21.32 & 8.06 &  & 36.14 & 3.64 & & 46.89 & 2.41 & & 40.01 & 2.37 & & 22.71 & 8.68 \\
\midrule
\texttt{Gemma-2-9B-It}       & 20.44 & 7.95 &  & 31.41 & 4.08 & & 43.47 & 2.66 & & 34.82 & 2.98 & & 26.05 & 6.66 \\
\texttt{LLaMA-3.1-8B-It}     & 11.38 & 14.42 &  & 25.36 & 5.88 & & 34.37 & 4.40 & & 29.05 & 3.96 & & 12.94 & 13.18 \\
\texttt{Command-R7B}         & 5.52 & 17.66 &  & 17.75 & 8.86 & & 16.51 & 10.98 & & 15.92 & 8.86 & & 8.72 & 16.50 \\
\texttt{Aya-expanse-32B}     & 12.01 & 13.60 &  & 28.25 & 4.64 & & 28.22 & 5.89 & & 28.70 & 4.00 & & 14.90 & 11.56 \\
\texttt{Aya-expanse-8B}      & 8.44 & 16.54 &  & 18.89 & 7.01 & & 12.75 & 11.65 & & 17.97 & 7.55 & & 8.66 & 15.76 \\
\texttt{Qwen-2.5-32B-It}     & 10.53 & 15.31 &  & 28.98 & 5.40 & & 28.98 & 6.67 & & 31.00 & 3.93 & & 15.93 & 12.47 \\
\texttt{Qwen-2.5-7B-It}      & 8.05 & 18.09 &  & 21.69 & 7.74 & & 14.79 & 12.09 & & 22.43 & 6.37 & & 9.54 & 16.22 \\
\midrule
\texttt{LLaMAX2-7B Alpaca}   & 12.35 & 11.21 &  & 21.17 & 8.13 & & 27.67 & 5.96 & & 12.71 & 13.89 & & 18.75 & 7.29 \\
\texttt{LLaMAX3-8B Alpaca}   & 24.03 & 5.57 &  & 31.00 & 4.02 & & 38.65 & 3.03 & & 32.03 & 3.30 & & 28.42 & 5.06 \\
\midrule
\texttt{LLaMA-2-7B}  \small{\textsc{5-shot Bm25}}        & 6.02 & 18.73 &  & 16.30 & 8.76 & & 10.78 & 12.84 & & 10.93 & 11.21 & & 8.17 & 16.63 \\
\texttt{LLaMA-3-8B}  \small{\textsc{5-shot Bm25}}        & 14.09 & 11.42 &  & 33.77 & 3.83 & & 39.46 & 3.41 & & 33.30 & 3.41 & & 17.08 & 10.34 \\
\midrule
\texttt{LLaMA-2-7B} uni.    & 15.94 & 11.93 &  & 30.78 & 6.44 & & 27.55 & 7.99 & & 21.30 & 8.17 & & 20.73 & 11.28 \\
\texttt{LLaMA-2-7B} uni.  \small{\textit{beam size=5}}   & 16.33 & 11.51 &  & 31.30 & 6.07 & & 28.77 & 7.44 & & 23.33 & 7.29 & & 19.68 & 10.47 \\
\texttt{LLaMA-3-8B} uni.    & 24.69 & 8.32 &  & 38.46 & 4.19 & & 42.77 & 3.78 & & 35.61 & 3.74 & & 31.38 & 6.98 \\
\texttt{LLaMA-3-8B} uni.  \small{\textit{beam size=5}}  & \bf 25.75 & \bf 7.87 &  & \bf 39.15 & \bf 3.98 & & \bf 44.04 & \bf 3.63 & & \bf 36.62 & \bf 3.51 & & \bf 32.14 & \bf 6.71 \\
\texttt{LLaMA-3-8B} multi.   & 23.29 & 9.05 &  & 37.55 & 4.49 & & 42.51 & 3.95 & & 35.44 & 3.91 & & 29.34 & 7.83 \\
\texttt{LLaMA-3-8B} multi. \small{\textit{beam size=5}}  & 24.53 & 8.63 &  & 38.53 & 4.34 & & 43.86 & 3.71 & & 36.33 & 3.69 & & 30.62 & 7.39 \\
\bottomrule
\end{tabular}
}
\end{center}
\caption{BLEU and MetricX scores for 10 X $\rightarrow$ English directions from FLORES~200\iffalse~\citep{goyal-etal-2022-flores, nllb2022}\fi. Best results after fine-tuning \iffalse (including any results that are not statistically worse)\fi are highlighted in bold.}
\label{tab:flores200_reverse}
\end{table*}
\subsection{ChrF++ and XCOMET scores} \label{appendix:chrf_and_xcomet}

We consider the XCOMET metric~\citep{guerreiro-etal-2024-xcomet}, specifically its reference-based version: \texttt{XCOMET-XL} (which supports the same 100 languages as XLM~RoBERTa~\citep{conneau-etal-2020-unsupervised}). XCOMET scores range from 0 and 1, which we rescale to 0 to 100 (higher scores are better). We evaluate the translations produced on FLORES-200 by models fine-tuned on the \textsc{TopXGen} dataset and report the results in Table~\ref{tab:flores200_chrf_xcomet}. The results are consistent with BLEU and MetricX.

\begin{table*}[ht]
\vskip 0.15in
\small
\begin{center}
\resizebox{\textwidth}{!}{
\begin{tabular}{lrrrrrrrrrrrrrr}
\toprule
\multirow{2}{*}{Methods}  & \multicolumn{2}{c}{Basque} & & \multicolumn{2}{c}{Hausa} & & \multicolumn{2}{c}{Igbo} & & \multicolumn{2}{c}{Kinyarwanda} & & \multicolumn{2}{c}{Nepali}\\
\cmidrule{2-3} \cmidrule{5-6} \cmidrule{8-9} \cmidrule{11-12} \cmidrule{14-15}
{} & {chrF++} & {XCOMET} &  & {chrF++} & {XCOMET} & & {chrF++} & {XCOMET} & & {chrF++} & {XCOMET} & & {chrF++} & {XCOMET}\\
\midrule
\texttt{NLLB-200-3.3B}       & 46.31 & 76.88 &  & 51.58 & 65.65 & & 40.56 & 19.04 & & 46.90 & 26.93 & & 44.59 & 75.82 \\
\texttt{Gemma-3-27B-It}      & 49.71 & 78.97 &  & 44.94 & 58.57 & & 36.41 & 18.40 & & 37.18 & 23.12 & & 45.76 & 83.12 \\
\midrule
\texttt{Gemma-2-27B-It}      & 46.69 & 73.43 &  & 42.23 & 58.72 & & 30.65 & 18.95 & & 28.31 & 24.61 & & 43.07 & 79.71 \\
\texttt{LLaMA-3.1-70B It}    & 49.28 & 77.61 &  & 43.40 & 55.46 & & 36.13 & 18.74 & & 29.38 & 24.69 & & 45.74 & 78.20 \\
\midrule
\texttt{Gemma-2-9B-It}       & 41.00 & 60.10 &  & 38.41 & 50.13 & & 25.82 & 19.76 & & 21.99 & 24.25 & & 39.87 & 74.51 \\
\texttt{LLaMA-3.1-8B-It}     & 39.97 & 51.19 &  & 31.30 & 31.23 & & 22.76 & 19.77 & & 19.25 & 23.33 & & 32.68 & 55.28 \\
\texttt{Command-R7B}         & 18.76 & 54.79 &  & 11.24 & 27.13 & & 9.60 & 23.63 & & 12.54 & 22.83 & & 23.29 & 50.46 \\
\texttt{Aya-expanse-32B}     & 32.58 & 30.74 &  & 23.35 & 21.84 & & 19.43 & 19.58 & & 22.10 & 24.56 & & 30.64 & 51.95 \\
\texttt{Aya-expanse-8B}      & 25.53 & 26.96 &  & 19.29 & 22.00 & & 17.22 & 19.02 & & 19.13 & 23.35 & & 22.01 & 62.55 \\
\texttt{Qwen-2.5-32B-It}     & 30.99 & 27.15 &  & 20.99 & 22.32 & & 20.45 & 20.23 & & 20.03 & 24.86 & & 30.76 & 45.73 \\
\texttt{Qwen-2.5-7B-It}      & 26.31 & 21.13 &  & 20.02 & 20.42 & & 15.90 & 19.98 & & 15.56 & 21.85 & & 23.11 & 31.12 \\
\midrule
\texttt{LLaMAX2-7B Alpaca}   & 31.16 & 34.21 &  & 40.75 & 36.32 & & 30.27 & 18.74 & & 20.11 & 24.34 & & 36.55 & 39.17 \\
\texttt{LLaMAX3-8B Alpaca}   & 34.38 & 56.20 &  & 41.54 & 55.12 & & 33.60 & 19.42 & & 19.07 & 25.07 & & 41.82 & 69.42 \\
\midrule
\texttt{LLaMA-2-7B} \small{\textsc{5-shot Bm25}}         & 16.90 & 22.75 &  & 11.64 & 25.21 & & 11.25 & 21.95 & & 13.92 & 23.18 & & 15.40 & 26.13 \\
\texttt{LLaMA-3-8B} \small{\textsc{5-shot Bm25}}         & 41.36 & 62.92 &  & 34.05 & 39.59 & & 23.20 & 20.60 & & 18.37 & 23.94 & & 37.47 & 61.10 \\
\midrule
\texttt{LLaMA-2-7B} uni.    & 36.44 & 39.97 &  & 37.46 & 41.80 & & 31.02 & 18.47 & & 26.53 & 22.55 & & 35.84 & 51.37 \\
\texttt{LLaMA-2-7B} uni.  \small{\textit{beam size=5}}   & 38.59 & 48.68 &  & 37.62 & 46.40 & & 32.41 & 17.56 & & 27.47 & 22.87 & & 38.38 & 61.84 \\
\texttt{LLaMA-3-8B} uni.   & 46.83 & 74.21 &  & 44.73 & 61.11 & & 36.65 & 18.53 & & 34.51 & 22.25 & & 42.99 & 78.17 \\
\texttt{LLaMA-3-8B} uni.  \small{\textit{beam size=5}} & \bf 48.28 & \bf 80.10 &  & \bf 45.21 & \bf 64.56 & & \bf 37.73 & \bf 18.60 & & \bf 36.32 & \bf 23.22 & & \bf 44.72 & \bf 82.32 \\
\texttt{LLaMA-3-8B} multi.  & 45.36 & 71.05 &  & 42.80 & 57.24 & & 35.65 & 18.35 & & 30.30 & 21.93 & & 42.39 & 75.78 \\
\texttt{LLaMA-3-8B} multi. \small{\textit{beam size=5}} & 47.02 & 77.89 &  & 43.76 & 61.57 & & 36.61 & 18.24 & & 32.23 & 23.08 & & 44.42 & 81.35 \\
\end{tabular}
}
\resizebox{\textwidth}{!}{
\begin{tabular}{lrrrrrrrrrrrrrr}
\toprule
\multirow{2}{*}{Methods}  & \multicolumn{2}{c}{Somali} & & \multicolumn{2}{c}{Sundanese} & & \multicolumn{2}{c}{Swahili} & & \multicolumn{2}{c}{Urdu} & & \multicolumn{2}{c}{Xhosa}\\
\cmidrule{2-3} \cmidrule{5-6} \cmidrule{8-9} \cmidrule{11-12} \cmidrule{14-15}
{} & {chrF++} & {XCOMET} &  & {chrF++} & {XCOMET} & & {chrF++} & {XCOMET} & & {chrF++} & {XCOMET} & & {chrF++} & {XCOMET}\\
\midrule
\texttt{NLLB-200-3.3B}       & 41.58 & 63.09 &  & 45.28 & 66.86 & & 58.60 & 78.97 & & 47.17 & 70.44 & & 46.78 & 51.58 \\
\texttt{Gemma-3-27B-It}      & 39.02 & 57.19 &  & 42.95 & 70.17 & & 58.37 & 81.34 & & 44.61 & 77.74 & & 38.44 & 39.86 \\
\midrule
\texttt{Gemma-2-27B-It}      & 32.03 & 35.44 &  & 39.54 & 61.69 & & 58.13 & 82.22 & & 32.32 & 33.64 & & 41.68 & 71.66 \\
\texttt{LLaMA-3.1-70B It}    & 33.33 & 36.81 &  & 42.48 & 67.08 & & 57.52 & 78.73 & & 46.05 & 73.17 & & 28.32 & 29.61 \\
\midrule
\texttt{Gemma-2-9B-It}       & 27.81 & 27.45 &  & 37.63 & 58.40 & & 53.64 & 75.32 & & 38.20 & 62.87 & & 27.10 & 28.04 \\
\texttt{LLaMA-3.1-8B-It}     & 22.56 & 21.65 &  & 33.22 & 48.63 & & 45.53 & 56.04 & & 38.60 & 55.91 & & 18.93 & 23.06 \\
\texttt{Command-R7B}         & 12.23 & 27.42 &  & 30.79 & 73.38 & & 26.97 & 25.13 & & 19.81 & 31.53 & & 14.80 & 24.35 \\
\texttt{Aya-expanse-32B}     & 28.18 & 26.07 &  & 33.84 & 52.53 & & 33.71 & 27.80 & & 31.86 & 44.64 & & 23.70 & 24.36 \\
\texttt{Aya-expanse-8B}      & 24.87 & 22.92 &  & 33.46 & 82.88 & & 25.24 & 21.24 & & 27.02 & 37.71 & & 21.26 & 22.46 \\
\texttt{Qwen-2.5-32B-It}     & 23.56 & 21.52 &  & 31.17 & 35.60 & & 34.26 & 26.47 & & 30.87 & 39.42 & & 23.01 & 24.18 \\
\texttt{Qwen-2.5-7B-It}      & 21.24 & 20.82 &  & 26.60 & 34.35 & & 25.25 & 21.48 & & 24.03 & 26.01 & & 18.34 & 21.66 \\
\midrule
\texttt{LLaMAX2-7B Alpaca}   & 33.31 & 30.17 &  & 31.55 & 41.34 & & 49.18 & 51.16 & & 32.08 & 27.12 & & 34.42 & 28.34 \\
\texttt{LLaMAX3-8B Alpaca}   & 35.14 & 50.75 &  & 34.94 & 56.56 & & 50.83 & 70.09 & & 38.90 & 57.27 & & 33.91 & 36.14 \\
\midrule
\texttt{LLaMA-2-7B} \small{\textsc{5-shot Bm25}}         & 13.76 & 22.46 &  & 24.55 & 36.89 & & 15.59 & 23.45 & & 15.02 & 24.66 & & 12.44 & 23.61 \\
\texttt{LLaMA-3-8B} \small{\textsc{5-shot Bm25}}         & 20.97 & 24.82 &  & 38.27 & 52.82 & & 46.86 & 60.37 & & 35.91 & 54.45 & & 16.69 & 24.76 \\
\midrule
\texttt{LLaMA-2-7B} uni.    & 32.56 & 32.35 &  & 39.99 & 60.48 & & 44.42 & 46.61 & & 34.98 & 40.43 & & 31.05 & 29.19 \\
\texttt{LLaMA-2-7B} uni.  \small{\textit{beam size=5}}   & 33.33 & 35.75 &  & 41.90 & 67.52 & & 46.66 & 55.64 & & 37.23 & 51.21 & & 32.80 & 33.36 \\
\texttt{LLaMA-3-8B} uni.   & 38.34 & 52.08 &  & 42.90 & 70.50 & & 55.21 & 75.45 & & 42.37 & 69.20 & & 36.84 & 38.35 \\
\texttt{LLaMA-3-8B} uni.  \small{\textit{beam size=5}} & \bf 39.08 & \bf 56.78 &  & \bf 44.18 & \bf 74.00 & & \bf 57.01 & \bf 80.62 & & \bf 43.36 & \bf 73.92 & & \bf 38.52 & \bf 44.37 \\
\texttt{LLaMA-3-8B} multi.  & 36.82 & 47.22 &  & 42.00 & 68.25 & & 53.11 & 72.29 & & 41.38 & 66.97 & & 35.77 & 36.21 \\
\texttt{LLaMA-3-8B} multi. \small{\textit{beam size=5}}  & 37.66 & 52.61 &  & 43.27 & 72.62 & & 54.94 & 78.22 & & 42.77 & 72.59 & & 37.37 & 42.73 \\
\bottomrule
\end{tabular}
}
\end{center}
\caption{ChrF++ and XCOMET-XL scores for 10 English $\rightarrow$ X directions from FLORES~200\iffalse~\citep{goyal-etal-2022-flores, nllb2022}\fi. Best fine-tuning results \iffalse(including any results that are not statistically worse)\fi are highlighted in bold.}
\label{tab:flores200_chrf_xcomet}
\end{table*}

\subsection{Results on NTREX-128 and TICO-19} \label{appendix:ntrex_and_tico}

In this section, we evaluate the models on 2 additional benchmarks:
\begin{itemize}[noitemsep, topsep=0pt, leftmargin=*]
    \item \textbf{NTREX~128}~\citep{barrault-etal-2019-findings, federmann-etal-2022-ntrex} is an MT benchmark derived from WMT19 news data translated by professional human translators. It contains 1997 parallel sentences and is recommended for the evaluation of from-English translation directions. We use the first 1000 sentence pairs for evaluation, and the last 997 sentence pairs as the selection pool.
    \item \textbf{TICO-19}~\citep{anastasopoulos-etal-2020-tico} is an MT benchmark comprising texts on the COVID-19 pandemic covering 35 languages. Its validation and test sets consist of 971 (used as a selection pool) and 2100 samples respectively.
\end{itemize}
We focus on translating from English. We report the results obtained on NTREX-19 in Table~\ref{tab:ntrex} and those obtained on TICO-19 in Table~\ref{tab:tico}. On both benchmarks, the best fine-tuned model usually ranks third, behind \texttt{NLLB-200-3.3B} and \texttt{Gemma-3-27B-It}.

\begin{table*}[ht]
\vskip 0.15in
\small
\begin{center}
\resizebox{\textwidth}{!}{
\begin{tabular}{lrrrrrrrrrrrrrr}
\toprule
\multirow{2}{*}{Methods}  & \multicolumn{2}{c}{Basque} & & \multicolumn{2}{c}{Hausa} & & \multicolumn{2}{c}{Igbo} & & \multicolumn{2}{c}{Kinyarwanda} & & \multicolumn{2}{c}{Nepali}\\
\cmidrule{2-3} \cmidrule{5-6} \cmidrule{8-9} \cmidrule{11-12} \cmidrule{14-15}
{} & {BLEU} & {MetricX} &  & {BLEU} & {MetricX} & & {BLEU} & {MetricX} & & {BLEU} & {MetricX} & & {BLEU} & {MetricX}\\
\midrule
\texttt{NLLB-200-3.3B}       & 18.29 & 8.29 &  & 25.43 & 4.51 & & 22.43 & 5.15 & & 21.05 & 4.76 & & 22.12 & 5.32 \\
\texttt{Gemma-3-27B-It}      & 21.62 & 5.79 &  & 19.54 & 4.97 & & 16.64 & 7.76 & & 14.98 & 7.99 & & 18.38 & 4.23 \\
\midrule
\texttt{Gemma-2-27B-It}      & 18.31 & 7.36 &  & 17.37 & 5.69 & & 14.07 & 10.44 & & 8.06 & 14.85 & & 17.39 & 4.81 \\
\texttt{LLaMA-3.1-70B It}    & 21.70 & 6.61 &  & 18.58 & 6.31 & & 18.44 & 8.76 & & 8.95 & 14.04 & & 18.42 & 5.15 \\
\midrule
\texttt{Gemma-2-9B-It}       & 13.40 & 10.14 &  & 14.69 & 6.94 & & 11.76 & 14.20 & & 5.03 & 19.88 & & 14.64 & 5.62 \\
\texttt{LLaMA-3.1-8B-It}     & 13.23 & 12.47 &  & 10.46 & 11.72 & & 10.07 & 17.09 & & 4.55 & 20.45 & & 9.60 & 9.09 \\
\texttt{Command-R7B}         & 4.02 & 13.21 &  & 2.56 & 18.52 & & 2.98 & 20.01 & & 2.92 & 21.52 & & 4.35 & 9.79 \\
\texttt{Aya-expanse-32B}     & 8.79 & 16.72 &  & 6.42 & 16.62 & & 5.65 & 20.08 & & 5.00 & 20.84 & & 8.02 & 9.08 \\
\texttt{Aya-expanse-8B}      & 5.04 & 19.93 &  & 4.86 & 16.32 & & 4.27 & 21.52 & & 3.29 & 22.82 & & 4.33 & 7.39 \\
\texttt{Qwen-2.5-32B-It}     & 6.91 & 19.06 &  & 6.25 & 16.65 & & 6.68 & 19.15 & & 4.23 & 22.44 & & 7.97 & 10.57 \\
\texttt{Qwen-2.5-7B-It}      & 4.69 & 21.65 &  & 5.12 & 18.91 & & 5.40 & 21.84 & & 2.97 & 23.83 & & 4.00 & 13.34 \\
\midrule
\texttt{LLaMAX2-7B Alpaca}   & 10.29 & 15.88 &  & 18.45 & 6.98 & & 10.10 & 13.79 & & 5.52 & 18.36 & & 9.82 & 14.73 \\
\texttt{LLaMAX3-8B Alpaca}   & 11.37 & 11.33 &  & 17.37 & 6.41 & & 15.10 & 9.80 & & 5.20 & 19.01 & & 16.97 & 6.48 \\
\midrule
\texttt{LLaMA-2-7B} \small{\textsc{5-shot Bm25}}         & 4.22 & 22.20 &  & 1.96 & 21.23 & & 3.57 & 21.84 & & 3.51 & 22.34 & & 2.79 & 19.29 \\
\texttt{LLaMA-3-8B} \small{\textsc{5-shot Bm25}}         & 14.61 & 10.07 &  & 12.56 & 10.41 & & 14.03 & 15.19 & & 5.45 & 20.42 & & 12.07 & 8.17 \\
\midrule
\texttt{LLaMA-2-7B} uni.    & 10.95 & 16.01 &  & 13.27 & 9.32 & & 13.01 & 11.64 & & 7.42 & 16.93 & & 10.51 & 9.51 \\
\texttt{LLaMA-2-7B} uni. \small{\textit{beam size=5}}   & 12.01 & 14.17 &  & 13.45 & 9.30 & & 14.35 & 10.70 & & 8.06 & 15.56 & & 11.62 & 7.96 \\
\texttt{LLaMA-3-8B} uni.    & 19.10 & 7.86 &  & 19.44 & 5.84 & & 17.88 & 8.19 & & 11.89 & 11.18 & & 15.89 & 5.36 \\
\texttt{LLaMA-3-8B} uni. \small{\textit{beam size=5}}   & \bf 20.77 & \bf 6.78 &  & \bf 20.02 & \bf 5.84 & & \bf 19.00 & \bf 7.38 & & \bf 13.16 & \bf 9.66 & & \bf 17.22 & \bf 4.81 \\
\texttt{LLaMA-3-8B} multi.   & 18.11 & 8.44 &  & 17.91 & 6.76 & & 16.60 & 8.55 & & 9.38 & 13.98 & & 15.35 & 5.79 \\
\texttt{LLaMA-3-8B} multi. \small{\textit{beam size=5}}  & 18.79 & 7.13 &  & 18.54 & 6.45 & & 18.18 & 8.04 & & 10.64 & 11.97 & & 16.10 & 5.00 \\
\end{tabular}
}
\resizebox{\textwidth}{!}{
\begin{tabular}{lrrrrrrrrrrrrrr}
\toprule
\multirow{2}{*}{Methods}  & \multicolumn{2}{c}{Somali} & & \multicolumn{2}{c}{Sundanese} & & \multicolumn{2}{c}{Swahili} & & \multicolumn{2}{c}{Urdu} & & \multicolumn{2}{c}{Xhosa}\\
\cmidrule{2-3} \cmidrule{5-6} \cmidrule{8-9} \cmidrule{11-12} \cmidrule{14-15}
{} & {BLEU} & {MetricX} &  & {BLEU} & {MetricX} & & {BLEU} & {MetricX} & & {BLEU} & {MetricX} & & {BLEU} & {MetricX}\\
\midrule
\texttt{NLLB-200-3.3B}       & 19.82 & 4.74 &  & - & - & & 38.63 & 5.38 & & 32.16 & 5.11 & & 18.73 & 4.61 \\
\texttt{Gemma-3-27B-It}      & 14.82 & 5.48 &  & - & - & & 36.97 & 4.23 & & 28.62 & 3.71 & & 12.41 & 8.58 \\
\midrule
\texttt{Gemma-2-27B-It}      & 11.04 & 10.83 &  & - & - & & 35.99 & 4.53 & & 26.89 & 4.49 & & 9.38 & 12.98 \\
\texttt{LLaMA-3.1-70B It}    & 12.03 & 10.82 &  & - & - & & 34.83 & 4.96 & & 30.50 & 4.24 & & 7.50 & 16.12 \\
\midrule
\texttt{Gemma-2-9B-It}       & 8.78 & 14.21 &  & - & - & & 32.52 & 5.93 & & 21.08 & 5.90 & & 7.61 & 18.14 \\
\texttt{LLaMA-3.1-8B-It}     & 5.79 & 19.23 &  & - & - & & 22.25 & 9.92 & & 21.19 & 6.70 & & 3.49 & 23.06 \\
\texttt{Command-R7B}         & 2.88 & 19.62 &  & - & - & & 6.90 & 19.57 & & 3.55 & 13.77 & & 2.44 & 21.81 \\
\texttt{Aya-expanse-32B}     & 8.53 & 14.47 &  & - & - & & 12.15 & 16.73 & & 11.18 & 8.44 & & 5.24 & 2.39 \\
\texttt{Aya-expanse-8B}      & 7.08 & 17.18 &  & - & - & & 6.81 & 21.40 & & 7.11 & 10.08 & & 4.03 & 23.73 \\
\texttt{Qwen-2.5-32B-It}     & 5.84 & 18.67 &  & - & - & & 12.29 & 17.44 & & 11.70 & 10.30 & & 4.98 & 22.21 \\
\texttt{Qwen-2.5-7B-It}      & 5.20 & 20.75 &  & - & - & & 6.86 & 21.38 & & 5.56 & 14.21 & & 3.00 & 24.10 \\
\midrule
\texttt{LLaMAX2-7B Alpaca}   & 11.38 & 9.57 &  & - & - & & 24.80 & 7.91 & & 11.74 & 15.73 & & 11.10 & 11.21 \\
\texttt{LLaMAX3-8B Alpaca}   & 13.66 & 7.30 &  & - & - & & 29.45 & 7.19 & & 22.58 & 6.60 & & 10.51 & 11.33 \\
\midrule
\texttt{LLaMA-2-7B} \small{\textsc{5-shot Bm25}}         & 2.78 & 21.07 &  & - & - & & 4.02 & 21.96 & & 3.62 & 19.31 & & 2.68 & 23.02 \\
\texttt{LLaMA-3-8B} \small{\textsc{5-shot Bm25}}        & 6.03 & 17.96 &  & - & - & & 26.37 & 8.55 & & 19.42 & 7.57 & & 3.25 & 21.78 \\
\midrule
\texttt{LLaMA-2-7B} uni.    & 10.92 & 12.75 &  & - & - & & 19.89 & 13.11 & & 14.05 & 9.78 & & 9.00 & 16.50 \\
\texttt{LLaMA-2-7B} uni. \small{\textit{beam size=5}}   & 11.18 & 11.31 &  & - & - & & 21.88 & 11.21 & & 15.42 & 8.20 & & 7.75 & 14.29 \\
\texttt{LLaMA-3-8B} uni.    & 15.30 & 7.77 &  & - & - & & 32.06 & 6.42 & & 22.34 & 5.25 & & 11.53 & 11.41 \\
\texttt{LLaMA-3-8B} uni. \small{\textit{beam size=5}}   & \bf 16.15 & \bf 6.83 &  & - & - & & \bf 34.31 & \bf 5.53 & & \bf 24.26 & \bf 4.62 & & \bf 12.18 & \bf 9.66 \\
\texttt{LLaMA-3-8B} multi.   & 14.04 & 8.74 &  & - & - & & 30.25  & 7.10 & & 21.80 & 5.69 & & 10.65 & 12.16 \\
\texttt{LLaMA-3-8B} multi. \small{\textit{beam size=5}}  & 14.79 & 7.80 &  & - & - & & 33.00 & 5.99 & & 23.51 & 4.97 & & 11.42 & 10.31 \\
\bottomrule
\end{tabular}
}
\end{center}
\caption{BLEU and MetricX scores for 9 English $\rightarrow$ X directions from NTREX~128~\citep{federmann-etal-2022-ntrex}. \iffalse Best results (including any results that are not statistically worse) are highlighted in bold.\fi}
\label{tab:ntrex}
\end{table*}

\begin{table*}[ht]
\vskip 0.15in
\small
\begin{center}
\resizebox{\textwidth}{!}{
\begin{tabular}{lrrrrrrrrrrrrrr}
\toprule
\multirow{2}{*}{Methods}  & \multicolumn{2}{c}{Basque} & & \multicolumn{2}{c}{Hausa} & & \multicolumn{2}{c}{Igbo} & & \multicolumn{2}{c}{Kinyarwanda} & & \multicolumn{2}{c}{Nepali}\\
\cmidrule{2-3} \cmidrule{5-6} \cmidrule{8-9} \cmidrule{11-12} \cmidrule{14-15}
{} & {BLEU} & {MetricX} &  & {BLEU} & {MetricX} & & {BLEU} & {MetricX} & & {BLEU} & {MetricX} & & {BLEU} & {MetricX}\\
\midrule
\texttt{NLLB-200-3.3B}       & - & - &  & 30.96 & 3.97 & & - & - & & 22.52 & 7.44 & & 32.72 & 4.02 \\
\texttt{Gemma-3-27B-It}      & - & - &  & 19.59 & 4.53 & & - & - & & 11.29 & 10.84 & & 27.37 & 3.44 \\
\midrule
\texttt{Gemma-2-27B-It}      & - & - &  & 18.60 & 5.29 & & - & - & & 6.93 & 17.01 & & 26.11 & 3.96 \\
\texttt{LLaMA-3.1-70B It}    & - & - &  & 20.98 & 5.62 & & - & - & & 8.07 & 15.57 & & 26.17 & 4.20 \\
\midrule
\texttt{Gemma-2-9B-It}       & - & - &  & 16.06 & 6.60 & & - & - & & 4.33 & 21.08 & & 22.37 & 4.55 \\
\texttt{LLaMA-3.1-8B-It}     & - & - &  & 11.83 & 11.38 & & - & - & & 3.53 & 21.47 & & 14.75 & 7.29 \\
\texttt{Command-R7B}         & - & - &  & 3.22 & 18.86 & & - & - & & 2.14 & 22.57 & & 7.24 & 8.56 \\
\texttt{Aya-expanse-32B}     & - & - &  & 6.80 & 16.78 & & - & - & & 3.73 & 22.19 & & 11.79 & 7.53 \\
\texttt{Aya-expanse-8B}      & - & - &  & 5.38 & 15.80 & & - & - & & 2.79 & 23.29 & & 7.47 & 6.37 \\
\texttt{Qwen-2.5-32B-It}     & - & - &  & 7.52 & 16.45 & & - & - & & 3.58 & 23.14 & & 12.52 & 9.22 \\
\texttt{Qwen-2.5-7B-It}      & - & - &  & 5.89 & 18.88 & & - & - & & 2.29 & 24.38 & & 6.89 & 12.67 \\
\midrule
\texttt{LLaMAX2-7B Alpaca}   & - & - &  & 17.64 & 6.58 & & - & - & & 5.01 & 19.74 & & 13.83 & 14.89 \\
\texttt{LLaMAX3-8B Alpaca}   & - & - &  & 19.49 & 5.81 & & - & - & & 4.29 & 20.06 & & 24.10 & 5.20 \\
\midrule
\texttt{LLaMA-2-7B} \small{\textsc{5-shot Bm25}}        & - & - &  & 2.62 & 19.59 & & - & - & & 2.98 & 21.64 & & 6.63 & 17.19 \\
\texttt{LLaMA-3-8B}  \small{\textsc{5-shot Bm25}}         & - & - &  & 14.40 & 8.96 & & - & - & & 5.13 & 20.08 & & 24.48 & 6.00 \\
\midrule
\texttt{LLaMA-2-7B} uni.    & - & - &  & 14.24 & 8.68 & & - & - & & 5.32 & 18.62 & & 13.46 & 9.07 \\
\texttt{LLaMA-2-7B} uni. \small{\textit{beam size=5}}   & - & - &  & 14.95 & 8.27 & & - & - & & 6.16 & 17.27 & & 14.80 & 7.67 \\
\texttt{LLaMA-3-8B} uni.    & - & - &  & 20.37 & 5.50 & & - & - & & 9.10 & 13.05 & & 22.79 & 4.68 \\
\texttt{LLaMA-3-8B} uni. \small{\textit{beam size=5}}   & - & - &  & \bf 21.31 & \bf 5.17 & & - & - & & \bf 10.09 & \bf 11.64 & & \bf 24.10 & \bf 4.40 \\
\texttt{LLaMA-3-8B} multi.   & - & - &  & 19.33 & 6.17 & & - & - & & 7.69 & 15.61 & & 22.17 & 4.90 \\
\texttt{LLaMA-3-8B} multi. \small{\textit{beam size=5}}  & - & - &  & 20.04 & 5.88 & & - & - & & 8.66 & 14.04 & & 23.35 & 4.33 \\
\end{tabular}
}
\resizebox{\textwidth}{!}{
\begin{tabular}{lrrrrrrrrrrrrrr}
\toprule
\multirow{2}{*}{Methods}  & \multicolumn{2}{c}{Somali} & & \multicolumn{2}{c}{Sundanese} & & \multicolumn{2}{c}{Swahili} & & \multicolumn{2}{c}{Urdu} & & \multicolumn{2}{c}{Xhosa}\\
\cmidrule{2-3} \cmidrule{5-6} \cmidrule{8-9} \cmidrule{11-12} \cmidrule{14-15}
{} & {BLEU} & {MetricX} &  & {BLEU} & {MetricX} & & {BLEU} & {MetricX} & & {BLEU} & {MetricX} & & {BLEU} & {MetricX}\\
\midrule
\texttt{NLLB-200-3.3B}       & 12.43 & 12.73 &  & - & - & & 36.95 & 3.95 & & 32.87 & 3.59 & & - & - \\
\texttt{Gemma-3-27B-It}      & 9.12 & 13.33 &  & - & - & & 34.35 & 3.67 & & 28.43 & 3.18 & & - & - \\
\midrule
\texttt{Gemma-2-27B-It}      & 6.61 & 16.67 &  & - & - & & 34.48 & 3.87 & & 25.84 & 4.08 & & - & - \\
\texttt{LLaMA-3.1-70B It}    & 6.74 & 16.98 &  & - & - & & 33.59 & 4.14 & & 27.66 & 3.65 & & - & - \\
\midrule
\texttt{Gemma-2-9B-It}       & 5.21 & 18.66 &  & - & - & & 30.23 & 4.79 & & 22.33 & 4.96 & & - & - \\
\texttt{LLaMA-3.1-8B-It}     & 3.58 & 21.88 &  & - & - & & 20.39 & 9.34 & & 20.41 & 5.75 & & - & - \\
\texttt{Command-R7B}         & 1.65 & 22.02 &  & - & - & & 6.55 & 19.48 & & 5.39 & 12.83 & & - & - \\
\texttt{Aya-expanse-32B}     & 4.76 & 18.98 &  & - & - & & 12.02 & 16.93 & & 12.00 & 7.59 & & - & - \\
\texttt{Aya-expanse-8B}      & 4.04 & 20.73 &  & - & - & & 7.83 & 21.26 & & 8.98 & 8.87 & & - & - \\
\texttt{Qwen-2.5-32B-It}     & 3.91 & 21.30 &  & - & - & & 13.00 & 16.95 & & 12.66 & 9.64 & & - & - \\
\texttt{Qwen-2.5-7B-It}      & 3.27 & 22.34 &  & - & - & & 7.82 & 21.32 & & 7.02 & 13.49 & & - & - \\
\midrule
\texttt{LLaMAX2-7B Alpaca}   & 7.13 & 16.01 &  & - & - & & 22.46 & 7.27 & & 10.95 & 16.62 & & - & - \\
\texttt{LLaMAX3-8B Alpaca}   & 8.05 & 14.42 &  & - & - & & 28.03 & 6.00 & & 22.61 & 5.15 & & - & - \\
\midrule
\texttt{LLaMA-2-7B}          & 1.35 & 22.57 &  & - & - & & 5.29 & 20.11 & & 5.02 & 17.49 & & - & - \\
\texttt{LLaMA-3-8B}          & 1.89 & 21.75 &  & - & - & & 25.87 & 7.04 & & 21.58 & 6.15 & & - & - \\
\midrule
\texttt{LLaMA-2-7B} uni.    & 6.41 & 17.53 &  & - & - & & 20.33 & 11.35 & & 14.35 & 8.90 & & - & - \\
\texttt{LLaMA-2-7B} uni. \small{\textit{beam size=5}}   & 6.85 & 16.72 &  & - & - & & 22.51 & 9.36 & & 16.35 & 7.13 & & - & - \\
\texttt{LLaMA-3-8B} uni.    & 9.16 & 14.42 &  & - & - & & 31.28 & 5.42 & & 23.50 & 4.31 & & - & - \\
\texttt{LLaMA-3-8B} uni. \small{\textit{beam size=5}}   & \bf 9.59 & \bf 13.90 &  & - & - & & \bf 32.61 & \bf 4.76 & & \bf 24.39 & \bf 3.94 & & - & - \\
\texttt{LLaMA-3-8B} multi.   & 8.26 & 15.25 &  & - & - & & 29.25 & 6.01 & & 22.82 & 4.64 & & - & - \\
\texttt{LLaMA-3-8B} multi. \small{\textit{beam size=5}}  & 8.80 & 14.47 &  & - & - & & 30.84 & 5.33 & & 24.07 & 4.20 & & - & - \\
\bottomrule
\end{tabular}
}
\end{center}
\caption{BLEU and MetricX scores for 6 English $\rightarrow$ X directions from TICO-19~\citep{anastasopoulos-etal-2020-tico}. \iffalse Best results (including any results that are not statistically worse) are highlighted in bold.\fi Best results after fine-tuning are highlighted in bold.}
\label{tab:tico}
\end{table*}

\subsection{Fine-tuning the generator and the back-translator} \label{appendix:teachers}

\begin{figure*}
\begin{center}
    \includegraphics[width=\linewidth]{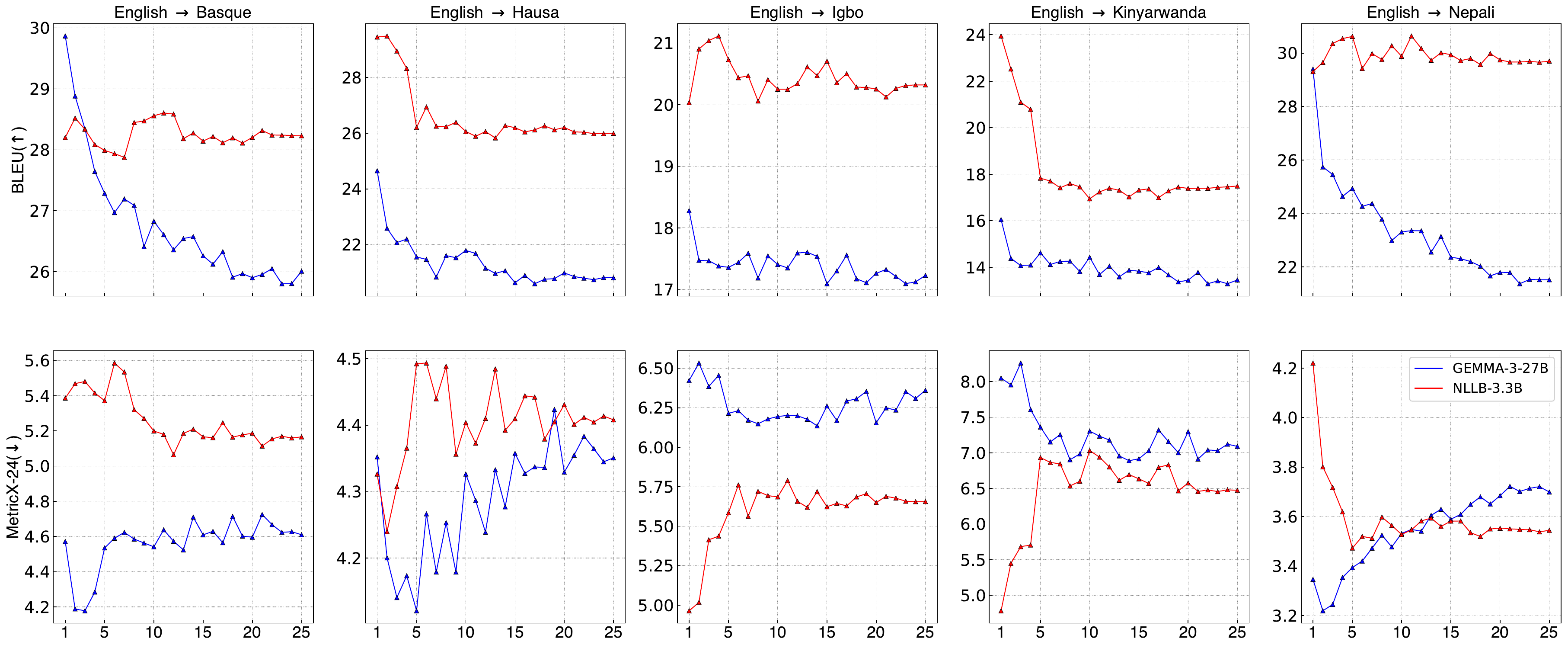}
    ~\\
    \includegraphics[width=\linewidth]{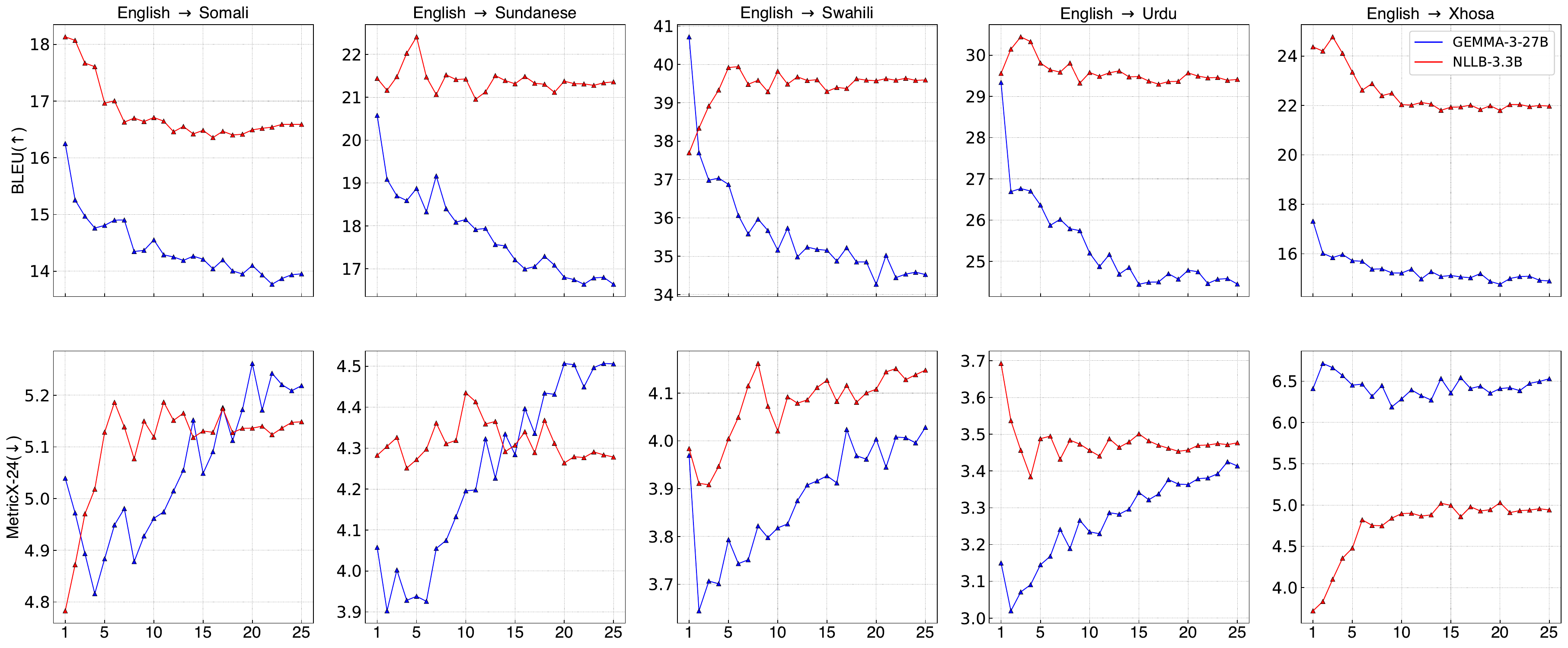}
    \caption{BLEU and MetricX results for 10 English$\rightarrow$X directions from FLORES~200. We fine-tune \texttt{NLLB-200-3.3B} and \texttt{Gemma-3-27B-PT}. We consider 1 model per direction and report the scores (greedy decoding) every 200 steps.}
    \label{fig:teachers}
\end{center}
\end{figure*}

We conduct additional experiments to assess the impact of fine-tuning both the back-translator, \texttt{NLLB-200-3.3B}, and the generator’s base model, \texttt{Gemma-3-27B-PT}, on the \textsc{TopXGen} dataset. For NLLB, we retain the same training hyperparameters as before, modifying only the maximum sequence length to 128. In contrast, for Gemma, we apply LoRA \citep{hu2022lora}, fine-tuning the \texttt{q\_proj}, \texttt{k\_proj}, \texttt{v\_proj}, \texttt{o\_proj}, \texttt{gate\_proj}, \texttt{up\_proj}, and \texttt{down\_proj} modules with rank $r=32$, scaling factor $\alpha=64$, and dropout rate 0.01. We use the same training hyperparameters as in previous runs and fine-tune on 4×H100 80GB GPUs. We compute BLEU and MetricX scores every 200 training steps and present the results in Figure~\ref{fig:teachers}. For languages like Basque, Nepali, and Urdu—where \texttt{Gemma-3-27B-IT} outperforms NLLB (see Table~\ref{tab:flores200})—we observe that fine-tuning NLLB on the \textsc{ToPXGen} dataset yields substantial improvements. Specifically, NLLB gains up to 3 BLEU points in Basque and Nepali, and 1 point in Urdu, with corresponding MetricX gains of approximately 1 point, and up to 3 in Nepali. In cases where NLLB and \texttt{Gemma-3-27B-IT} perform comparably (e.g., Sundanese and Swahili), BLEU remains largely unchanged, while MetricX shows modest improvements. Conversely, when \texttt{Gemma-3-27B-IT} underperforms NLLB, further fine-tuning NLLB on \textsc{ToPXGen} leads to a decline in performance—more pronounced in BLEU, but also noticeable in MetricX. Fine-tuning \texttt{Gemma-3-27B-PT} on generations from \texttt{Gemma-3-27B-IT} yields significant gains in translation quality. However, we observe diminishing returns as training progresses across all directions. We hypothesize that the model does not learn translation per se through this process; rather, early training steps help it infer the structure of the translation task. Once this template is internalized, the model relies on its pretrained knowledge for translation. Further fine-tuning on its own generations appears counterproductive.
As shown in Table~\ref{tab:small}, the fine-tuned models do not surpass the performance of \texttt{Gemma-3-27B-PT} in the \textsc{5-shot Bm25} setting, suggesting that self-generated data does not enhance the model’s understanding of the target languages.

\begin{table*}[ht]
\vskip 0.15in
\small
\begin{center}
\resizebox{\textwidth}{!}{
\begin{tabular}{lrrrrrrrrrrrrrr}
\toprule
\multirow{2}{*}{Methods}  & \multicolumn{2}{c}{Basque} & & \multicolumn{2}{c}{Hausa} & & \multicolumn{2}{c}{Igbo} & & \multicolumn{2}{c}{Kinyarwanda} & & \multicolumn{2}{c}{Nepali}\\
\cmidrule{2-3} \cmidrule{5-6} \cmidrule{8-9} \cmidrule{11-12} \cmidrule{14-15}
{} & {BLEU} & {MetricX} &  & {BLEU} & {MetricX} & & {BLEU} & {MetricX} & & {BLEU} & {MetricX} & & {BLEU} & {MetricX}\\
\midrule
\texttt{Gemma-3-27B-PT} \small{\textsc{5-shot Bm25}}       & 31.04 & 3.73 &  & 25.98 & 4.24 & & 19.39 & 6.10 & & 21.02 & 5.77 & & 34.06 & 3.71 \\
\texttt{Gemma-3-4B-PT}  \small{\textsc{5-shot Bm25}}       & 19.78 & 8.29 &  & 15.41 & 8.33 & & 10.15 & 13.73 & & 6.98 & 17.32 & & 25.48 & 5.18 \\
\texttt{Gemma-3-4B-It}         & 9.39 & 15.31 &  & 6.95 & 13.04 & & 5.81 & 18.72 & & 4.15 & 21.22 & & 16.99 & 6.56 \\
\midrule
\texttt{Gemma-3-4B-PT} uni.   & 22.57 & 6.44 &  & 19.22 & 5.29 & & 15.88 & 7.64 & & 11.35 & 10.70 & & 21.25 & 4.02 \\
\texttt{Gemma-3-4B-PT} multi.   & 21.94 & 6.93 &  & 18.12 & 6.05 & & 14.96 & 8.45 & & 9.57 & 13.42 & & 22.31 & 4.13 \\
\end{tabular}
}
\resizebox{\textwidth}{!}{
\begin{tabular}{lrrrrrrrrrrrrrr}
\toprule
\multirow{2}{*}{Methods}  & \multicolumn{2}{c}{Somali} & & \multicolumn{2}{c}{Sundanese} & & \multicolumn{2}{c}{Swahili} & & \multicolumn{2}{c}{Urdu} & & \multicolumn{2}{c}{Xhosa}\\
\cmidrule{2-3} \cmidrule{5-6} \cmidrule{8-9} \cmidrule{11-12} \cmidrule{14-15}
{} & {BLEU} & {MetricX} &  & {BLEU} & {MetricX} & & {BLEU} & {MetricX} & & {BLEU} & {MetricX} & & {BLEU} & {MetricX}\\
\midrule
\texttt{Gemma-3-27B-PT}   \small{\textsc{5-shot Bm25}}     & 16.99 & 4.96 &  & 24.54 & 3.88 & & 42.17 & 3.91 & & 29.72 & 3.33 & & 19.10 & 5.41 \\
\texttt{Gemma-3-4B-PT}    \small{\textsc{5-shot Bm25}}     & 9.77 & 10.42 &  & 16.91 & 7.41 & & 30.83 & 6.65 & & 21.32 & 4.97 & & 8.63 & 15.04 \\
\texttt{Gemma-3-4B-It}         & 4.93 & 15.94 &  & 9.32 & 8.68 & & 18.31 & 10.75 & & 16.16 & 6.33 & & 4.08 & 20.60 \\
\midrule
\texttt{Gemma-3-4B-PT} uni.   & 13.04 & 6.41 &  & 17.29 & 4.62 & & 32.00 & 5.10 & & 22.39 & 3.88 & & 12.60 & 9.24 \\
\texttt{Gemma-3-4B-PT} multi.   & 12.22 & 7.31 &  & 16.49 & 5.71 & & 30.39 & 5.51 & & 21.92 & 4.09 & & 11.64 & 10.50 \\
\bottomrule
\end{tabular}
}
\end{center}
\vskip -0.1in
\caption{BLEU and MetricX scores for 10 English $\rightarrow$ X directions from FLORES. Best results after fine-tuning are highlighted in bold.}
\label{tab:small}
\end{table*}

\subsection{Analysis of the \textsc{TopXGen} dataset} \label{appendix:dataset}
For each language, we compute the average \texttt{MetricX-24} quality estimation (QE) scores over the first 20K sentences. We also report the average number of words and tokens per sentence for both the source (English) and target sides. As shown in Table~\ref{tab:mean}, source sentences have a relatively consistent average word count across languages. However, in terms of tokens, sentences in low-resource languages (LRLs) typically require twice as many tokens as their English counterparts. For language pairs involving Hausa, Nepali, Somali, and Urdu, \textsc{TopXGen} achieves higher QE scores than both FLORES and SMOL, suggesting that its topic-guided generation produces natural and coherent text in LRLs, accurately translated by the back-translation model.
\\
The Vendi Score \citep{dan2023vendi}, calculated using SONAR embeddings, quantifies the diversity of a text—higher values indicate greater diversity. The results are summarized in Table~\ref{tab:mean}. On the target side, \textsc{TopXGen} generally achieves higher Vendi scores than FLORES (e.g., 1.123 vs. 1.096 in Somali), suggesting more diverse generations. Source-side sentences are also more diverse in \textsc{TopXGen}, though the difference is less pronounced. Notably, \textsc{SmolSent}, despite its smaller size, exhibits high diversity and occasionally surpasses \textsc{TopXGen}—particularly in languages like Hausa on the target side. However, as shown in Figure~\ref{fig:real}, this increased diversity does not consistently lead to better translation quality than that achieved by \textsc{TopXGen}.
\\
\begin{table*}[ht]
\captionsetup{skip=0pt}
\small
\begin{center}
\resizebox{\linewidth}{!}{
\begin{tabular}{lrrrrrrrrrr}
\toprule
{} & Basque & Hausa & Igbo & Kinyarwanda & Nepali & Somali & Sundanese & Swahili & Urdu & Xhosa \\
\midrule
Source mean number of words                             & 22.18 & 23.47 & 21.10 & 22.80 & 18.21 & 22.80 & 22.15 & 23.02 & 19.04 & 23.03 \\
Target mean number of words                             & 17.49 & 26.72 & 23.05 & 19.12 & 16.16 & 25.50 & 19.69 & 22.96 & 24.30 & 15.41 \\
Source mean \texttt{Gemma-3-27B-It} tokens              & 27.91 & 28.84 & 26.46 & 28.33 & 22.84 & 28.20 & 27.87 & 28.38 & 23.54 & 28.62 \\
Target mean \texttt{Gemma-3-27B-It} tokens              & 46.11 & 50.44 & 54.32 & 57.20 & 34.32 & 57.44 & 43.06 & 48.89 & 35.08 & 56.02 \\
\midrule
\texttt{MetricX24-XXL} QE scores \textsc{TopXGen}       & 3.19 & \bf 3.58 & 5.68 & 5.38 & \bf 2.86 & \bf 4.34 & 3.88 & 2.91 & \bf 2.62 & 5.89 \\
\texttt{MetricX24-XXL} QE scores FLORES                 & \bf 2.65 & 3.63 & 5.02 & \bf 3.43 & 4.13 & 4.89 & \bf 3.66 & 3.58 & 3.69 & \bf 3.82 \\
\texttt{MetricX24-XXL} QE scores SMOL                   & - & 4.00 & \bf 4.62 & 3.97 & - & 6.50 & - & 3.87 & - & 3.83 \\
\midrule
{} & \multicolumn{10}{c}{Vendi scores \citep{dan2023vendi} based on SONAR embeddings \citep{duquenne2023sonarsentencelevelmultimodallanguageagnostic}} \\
\midrule
\textsc{TopXGen} Source                  & 1.075 & 1.070 & 1.069 & 1.068 & 1.089 & 1.072 & 1.072 & 1.078 & 1.086 & 1.066 \\
\textsc{TopXGen} Target                  & 1.103 & 1.098 & 1.103 & 1.113 & 1.115 & 1.123 & 1.107 & 1.106 & 1.106 & 1.107 \\
\midrule
\textsc{SmolSent} Source                       & - & 1.076 & 1.076 & 1.076 & - & 1.076 & - & 1.076 & - & 1.076 \\
\textsc{SmolSent} Target                       & - & 1.125 & 1.116 & 1.106 & - & 1.134 & - & 1.114 & - & 1.113 \\
\midrule
\textsc{FLORES} Source                         & 1.051 & 1.051 & 1.051 & 1.051 & 1.051 & 1.051 & 1.051 & 1.051 & 1.051 & 1.051 \\
\textsc{FLORES} Target                         & 1.066 & 1.074 & 1.074 & 1.081 & 1.077 & 1.096 & 1.086 & 1.081 & 1.067 & 1.077 \\
\bottomrule
\end{tabular}
}
\end{center}
\caption{Statistics of the \textsc{TopXGen} dataset in comparison to FLORES and SMOL.}
\label{tab:mean}
\end{table*}
We also evaluate how well the paragraphs generated by \textsc{TopXGen} align with their intended topics. To do this, we generate 1,000 paragraphs per language using \texttt{Gemma-3-27B-It}, and ask both \texttt{Gemma-3-27B-It} and \texttt{Llama-4-Scout-17B-16E-Instruct} to assess whether each paragraph accurately addresses its assigned topic. We consider two settings: the paragraph written in the low-resource language, and its English translation obtained via sentence-by-sentence translation using \texttt{NLLB-3.3B}. Results are presented in Table~\ref{tab:topics}. According to \texttt{Gemma-3-27B-It}, 97\% of the paragraphs it generates are on-topic, though this rate decreases to 90\% when the same paragraphs are translated into English. Similarly, \texttt{Llama-4-Scout-17B-16E-Instruct} finds that 93\% of the original paragraphs align with their topics, dropping to 83–85\% after translation. In summary, the generated paragraphs are generally well-aligned with the provided topics. Even in cases where strict topical alignment is not achieved, the content remains relevant for machine translation training, where the primary requirement is having semantically equivalent sentences across languages.

\begin{table*}[ht]
\captionsetup{skip=0pt}
\small
\begin{center}
\resizebox{\linewidth}{!}{
\begin{tabular}{lrrrrrrrrrr}
\toprule
{} & Basque & Hausa & Igbo & Kinyarwanda & Nepali & Somali & Sundanese & Swahili & Urdu & Xhosa \\
\midrule
{} & \multicolumn{10}{c}{\texttt{Gemma-3-27B-It}} \\
\midrule
{} & \multicolumn{10}{c}{\textit{In English}} \\
\midrule
Yes                            & 972 & 973 & 982 & 965 & 980 & 973 & 965 & 977 & 977 & 969 \\
No                             &  28 &  27 &  18 &  35 &  20 &  27 & 35 &  23 &  23 &  31 \\
\midrule
{} & \multicolumn{10}{c}{\textit{In LRL}} \\
\midrule
Yes                            & 946 & 924 & 903 & 878 & 938 & 926 & 930 & 943 & 913 & 891 \\
No                             &  54 &  76 &  97 & 122 &  62 &  74 & 70 &  57 &  87 & 109 \\
\midrule 
{} & \multicolumn{10}{c}{\texttt{Llama-4-Scout-17B-16E-Instruct}} \\
\midrule
{} & \multicolumn{10}{c}{\textit{In English}} \\
\midrule
Yes                            & 904 & 935 & 903 & 919 & 956 & 934 & 918 & 902 & 959 & 919 \\
No                             & 59  &  53 &  77 & 77 &  23 &  63 &  82 &  67 &  38 &  70 \\
\midrule
{} & \multicolumn{10}{c}{\textit{In LRL}} \\
\midrule
Yes                            & 896 & 854 & 837 & 832 & 903 & 881 & 881 & 894 & 871 & 805 \\
No                             & 100 & 143 & 159 & 163 &  96 & 112 & 114 & 103 & 127 & 192 \\
\bottomrule
\end{tabular}
}
\end{center}
\caption{Repartition of 1000 paragraphs based on whether they discuss the topic they correspond to. We query \texttt{Gemma-3-27B-It} and \texttt{Llama-4-Scout-17B-16E-Instruct} \citep{MetaAI_LLaMA-4} in two settings: before and after translation by the back-translator (\texttt{NLLB-200--3.3B}).}
\label{tab:topics}
\end{table*}

\begin{table*}[ht]
    \centering\small
    \begin{center}
    \resizebox{\linewidth}{!}{
    \begin{tabular}{lll}
    \toprule
     Ground Truth Topics & \texttt{gpt-4o-mini-2024-07-18} Labels & Most Representative Words \\
    \midrule
     Theodore Shackley                                 & CIA interventions in Peru & cia | the | united | of | states | in | training| , | war | was \\
     Prasophyllum atratum                              & Biodiversity and conservation & species | the | australia | is | , | flowers | of | . | plant | it \\
     Frank Shields                                     & Development of Alternative Sports & tennis | sport | , | . | and | the | to | in | is | players \\
     Guardians of the Galaxy Vol. 3 (soundtrack)       & Influence of Music in Film & film | music | the | nosov | , | soundtrack | of | 's | a | . \\
     Llanfawr Quarries                                 & Local heritage and infrastructure & the | monuments | quarries | , | of | castle | and | . | these | to \\
     Deng Xi                                           & Influential Chinese Scholars & the | , | of | . | and | dynasty | chinese | qing | to | a \\
     \bottomrule
    \end{tabular}
    }
    \end{center}
    \vskip -0.1in
    \caption{Alignement between the ground truth topics and topics derived by BERTopic on six samples.}
    \label{tab:topic_modeling}
\end{table*}

\subsection{Topic Modeling} \label{appendix:topic_modeling}
We use BERTopic\footnote{\url{https://maartengr.github.io/BERTopic/index.html}} \citep{grootendorst2022bertopic} to assess whether the most relevant words identified through clustering align with the intended topics. As shown in Table~\ref{tab:topic_modeling}, we present results on six Basque paragraphs translated into English from the \textsc{TopXGen} dataset. We experiment with two setups: (1) using \texttt{gpt-4o-mini} to generate a topic label for each paragraph, and (2) extracting the top 10 words most relevant to each paragraph’s context. We find that the GPT-generated topics often encompass the ground truth topics, which are typically more fine-grained. For instance, Theodore Shackley is identified as a CIA officer, Frank Shields as a tennis player, and Deng Xi as a Chinese philosopher-the relevant words reflect these identities accurately.


\subsection{Qualitative Analysis} \label{appendix:qualitative_analysis}

In Table~\ref{tab:generations}, we observe that \texttt{Gemma-3-27B-It}’s generations in share lexical overlap—both at the character and word level—with Google Translate's translations of the corresponding English sentences, as identified by the back-translator. Furthermore, Google Translate consistently produces translations that are similar to those obtained from back-translating \texttt{Gemma-3-27B-It}’s generations. This suggests that the model is not hallucinating content, but instead generating semantically faithful and plausible sentences in the target language, reinforcing the relevance of our data generation pipeline.
\\
\begin{table*}[ht]
\centering\small
\begin{tabular}{p{7.5cm}p{7.5cm}}
\toprule
\multicolumn{2}{c}{Basque}\\
\midrule
Hauetako bakoitzak bere ezaugarriak ditu, batzuk produktu freskoen saldaritzan espezializatuz (okindegiak, frutariak) eta beste batzuk otoitz-zerbitzu ezarritik haratago doazen produktu eta zerbitzuak eskainiz (oinarrizko janari-elementuen salmenta nagusitzen duten tabako-dendetan, adibidez). & Each of these has its own characteristics, some specializing in the sale of fresh products (farms, fruit shops) and others offering products and services that go beyond the established prayer service (for example in tobacco shops that dominate the sale of basic food items). \\
\textcolor{red}{Hauetako bakoitzak bere ezaugarriak ditu, batzuk produktu freskoen salmentan espezializatuta daude (baserriak, fruta-dendak) eta beste batzuk otoitz-zerbitzu finkatutik haratago doazen produktuak eta zerbitzuak eskaintzen dituzte (adibidez, oinarrizko elikagaien salmentan nagusi diren tabako-dendetan).} & \textcolor{blue}{Each of these has its own characteristics, with some specializing in the sale of fresh produce (bakeries, greengrocers) and others offering products and services that go beyond the established prayer service (for example, tobacconists who mainly sell basic food items).}\\
\midrule
\multicolumn{2}{c}{Hausa}\\
\midrule
Wadannan matakan sun hada da sabbin kayayyakin safarar dukiya, da kuma inganta tsaron filin yayin da ake tafiya da komo. & These measures included new means of transporting goods, as well as improved field security during transportation.\\
\textcolor{red}{Wadannan matakan sun hada da sabbin hanyoyin jigilar kayayyaki, da kuma inganta tsaro a fagen sufuri.} & \textcolor{blue}{These measures include new transportation equipment, and improved field security during travel and return.} \\
\midrule
\multicolumn{2}{c}{Kinyarwanda}\\
\midrule
Impamvu Korea Times yifashishwa cyane, ni uko idashyira agahato ku makuru, kandi ngo ikunda kugaragaza ibintu bitandukanye na byinshi bisanzwe bimenyeshwa n’izindi nzego za Leta. & The reason for the popularity of the Korea Times is that it does not censor information, and tends to present information that differs from most other government agencies. \\
\textcolor{red}{Impamvu yo kwamamara muri Korea Times nuko idakurikirana amakuru, kandi ikunda kwerekana amakuru atandukanye nizindi nzego za leta.} & \textcolor{blue}{The reason the Korea Times is so widely used is that it does not impose restrictions on news, and it tends to present things that are different from what is usually reported by other government agencies.} \\
\midrule
\multicolumn{2}{c}{Somali}\\
\midrule
Qoyskiisu waxay ahaayeen kuwo qani ah oo leh xiriirro badan, taasoo ka caawisay inuu helaa fursado badan oo uu kaga shaqeeyo adeegga milatari. & His family was wealthy and had many connections, which helped him get many opportunities to work in the military service.\\
\textcolor{red}{Qoyskiisu waxay ahaayeen kuwo hodan ah oo lahaa xidhiidho badan, taas oo ka caawisay inuu helo fursado badan oo uu kaga shaqeeyo adeegga milatariga.} & \textcolor{blue}{His family was wealthy and well-connected, which helped him find many opportunities to serve in the military.} \\
\midrule
\multicolumn{2}{c}{Sundanese}\\
\midrule
Inskripsi ieu, ditulis dina basa Latin jeung basa Yunani, mangrupa conto anu saé kana kabijakan administrasi Romawi anu ngamimitahan panggunaan basa lokal pikeun mastikeun komunikasi anu efektif jeung populasi setempat. & The inscription, written in Latin and Greek, is a good example of a Roman administrative policy that encouraged the use of local languages to ensure effective communication with the local population.\\
\textcolor{red}{Prasasti, ditulis dina basa Latin sarta Yunani, mangrupakeun conto alus ngeunaan kawijakan administrasi Romawi nu wanti pamakéan basa lokal pikeun mastikeun komunikasi éféktif jeung populasi lokal.} & \textcolor{blue}{This inscription, written in Latin and Greek, is a good example of the Roman administrative policy of encouraging the use of local languages to ensure effective communication with the local population.} \\
\midrule
\multicolumn{2}{c}{Swahili}\\
\midrule
Pamoja na uzuri wake wa pekee, mlima huu pia umekuwa chanzo cha hadithi na misemo ya kitaifa kwa watu wa Wales kwa muda mrefu. & Along with its unique beauty, this mountain has also long been the source of stories and national sayings for the people of Wales. \\
\textcolor{red}{Pamoja na uzuri wake wa kipekee, mlima huu pia kwa muda mrefu umekuwa chanzo cha hadithi na misemo ya kitaifa kwa watu wa Wales} & \textcolor{blue}{Along with its unique beauty, this mountain has also long been the source of national legends and sayings for the Welsh people.} \\
\midrule
\multicolumn{2}{c}{Xhosa}\\
\midrule
Kukho imilinganiselo eyahlukeneyo esetyenziswa ukunje, ngokuphumela kwindlela yokuxabisa iimpahla, kodwa kwakungekho zibakala ezivela kunyaka. & Various measurements are used, resulting in a valuation system, but there were no facts from the year. \\
\textcolor{red}{Imilinganiselo eyahlukeneyo iyasetyenziswa, ekhokelela kwinkqubo yokuxabisa, kodwa akuzange kubekho zibakala zonyaka.} & \textcolor{blue}{There are different measures used today, resulting in a way of valuing goods, but there were no facts from the year.} \\
\bottomrule
\end{tabular}
\caption{Examples of generations with their back-translator's translations\iffalse.\fi. We provide in red, Google Translate's translations (in the source language) of \texttt{NLLB-200-3.3B}'s translations in English. We provide in blue, Google Translate's translations (in English) of the generator's (\texttt{Gemma-3-27B-It}) generations.}
\label{tab:generations}
\end{table*}

\end{document}